\documentclass[a4paper,12pt]{report}
\usepackage[utf8]{inputenc}
\usepackage[top=1.4in, bottom=1.3in, left=1.3in, right=1.3in]{geometry} 
\usepackage{amsmath} 
\usepackage{amsfonts}
\usepackage{graphicx} 
\usepackage{titling} 
\thanksmarkseries{fnsymbol} 

\usepackage{setspace} 
\usepackage{listings} 
\usepackage{url} 
\usepackage{caption}
\usepackage{subcaption} 
\usepackage{pdflscape} 
\usepackage{pdfpages} 

\usepackage{amsthm} 

\newtheorem{definition}{Definition}[section]

\newtheorem{proposition}{Proposition}[section]

\usepackage[ruled]{algorithm2e}
\usepackage{algorithmic}

\usepackage{adjustbox}

\usepackage{titlesec}
\titleformat{\chapter}[display]{\normalfont\huge\bfseries}{\chaptertitlename\ \thechapter}{13pt}{\Huge}
\titlespacing*{\chapter}{0pt}{10pt}{18pt}

\newcommand\independent{\protect\mathpalette{\protect\independenT}{\perp}}
\def\independenT#1#2{\mathrel{\rlap{$#1#2$}\mkern2mu{#1#2}_p}} 

\usepackage{breqn}

\usepackage{makecell}

\usepackage{appendix}

\usepackage{hyperref} 
\hypersetup{
    pdftex, 
    linktoc=all 
}

\usepackage[backend=biber]{biblatex} 
\bibliography{bn-structure-ilp.bib}

\title{Solving Bayesian Network Structure Learning Problem with Integer Linear Programming}
\author{Ronald Seoh}
\date{01 Sep 2015}

\begin{document}

\begin{titlepage}

\thispagestyle{empty}

\begin{center}

\begin{minipage}{\textwidth}
    \centering
    \vspace{2cm}
    {\Large \textbf{Solving Bayesian Network Structure Learning Problem with Integer Linear Programming}\par}
    \vspace{5cm}
    {\large Ronald Seoh\par}
    \vspace{5cm}
    {\large A Dissertation Submitted to the Department of Management \\of the London School of Economics and Political Science \\for the Degree of Master of Science\par}
    \vspace{5cm}
    {\large 01 Sep 2015}
\end{minipage}

\end{center}

\end{titlepage}

\pagenumbering{gobble}

\subsection*{Abstract}

This dissertation investigates integer linear programming (ILP) formulation of Bayesian Network structure learning problem. We review the definition and key properties of Bayesian network and explain score metrics used to measure how well certain Bayesian network structure fits the dataset. We outline the integer linear programming formulation based on the decomposability of score metrics.

In order to ensure acyclicity of the structure, we add ``cluster constraints'' developed specifically for Bayesian network, in addition to cycle constraints applicable to directed acyclic graphs in general. Since there would be exponential number of these constraints if we specify them fully, we explain the methods to add them as cutting planes without declaring them all in the initial model. Also, we develop a heuristic algorithm that finds a feasible solution based on the idea of sink node on directed acyclic graphs.

We implemented the ILP formulation and cutting planes as a \textsf{Python} package, and present the results of experiments with different settings on reference datasets.




\tableofcontents

\pagenumbering{arabic}

\chapter{Introduction}

Bayesian network is a probabilistic graphical model using directed acyclic graphs to express joint probability distributions and conditional dependencies between different random variables. Nodes represent random variables and directed arcs are drawn from parent nodes to child nodes to show that the child node is conditionally dependent to its parent nodes. Aside from its mathematical properties, Bayesian network's visual presentation make them easily perceivable, and many researchers in different fields have used it to model and study their systems.

Constructing a Bayesian network requires two major components: its graph topology, and parameters for the joint probability distribution. In some cases, the structure of the graph gets gets specified in advance by ``experts" and we find the values for the parameters that fit the given data, more specifically by using maximum likelihood approach. 

The problem gets more complicated when we do not know the graph structure and have to learn it from the given data. This would be the case when the problem domain is too large and it is extremely difficult or impractical for humans to manually define the structure. Learning the structure of the network from data have been proven NP-hard, and there have been several different approaches over the years to tackle this problem.

Early methods were approximate searches, where the algorithm searches through the candidates to look for the most probable one based on their strategies but usually does not provide any guarantee of optimality. Later on, there have developments on exact searches based on conditional independence testing or dynamic programming. While these did provide some level of optimality, their real world applicability was limited as the amount of computation became infeasible for larger number of datasets, or the underlying assumptions could not be easily met in reality. 

In this dissertation, we discuss another method of conducting exact search of Bayesian network structure using integer linear programming (ILP). While this approach is relatively new in the field, it achieves fast search process based on various integer programming techniques and state-of-the-art solvers, and allows users to incorporate prior knowledge easily as constraints. 

One of the main challenges in Bayesian network structure learning is on how to enforce acyclicity of the resulting structure. While acyclicity constraint developed for general DAGs also applies to Bayesian network, it is not tight enough for our ILP formulation due to the fact that we select the set of parent nodes for each variables rather than individual edges in the graph. We study so-called ``cluster constraints'' developed by Jaakkola et al.\autocite{jaakkola2010learning} that provides stronger enforcement of acyclicity on Bayesian network structure. Since there will be exponential number of such constraints if we specify them fully, we also cover how we can add them as cutting planes when needed during the solution process.

We also implemented a test computer program based on this ILP formulation and examined their performance on some of the sample datasets. There are few other implementation publicily available, but we provide a clean object-oriented version written in \textsf{Python} programming language.

The structure of this disseration will be as follows: Chapter 2 will review the concept of integer linear programming and Bayesian network needed to understand the problem. Chapter 3 will examine score metrics used for BN structures, and present the ILP formulation. Chapter 4 will explain our approach on adding cluster constraints as cutting planes to the ILP model, and a heuristic algorithm to obtain good feasible solutions. Chapter 5 will present our implementation details and benchmark results. Lastly, Chapter 6 will provide pointers to further development of our methods.

\chapter{Preliminaries}
This chapter reviews two theoretical and conceptual foundations behind the topic of this dissertation: Bayesian Network (BN) and Integer Linear Programming (ILP).

\section{Bayesian Network}

\subsection{Overview}

Consider a dataset \(D\) that contains \(n\) predictor variables \(x_1, x_2, x_3, ... , x_n\), the class variable \(y\) that can take \(m\) classes \(y_1, y_2, y_3, ... , y_m\) and \(k\) samples \(s_1, s_2, s_3, ... , s_k\). Let's suppose we want to find probabilities for each possible values of \(y\), given the observations of all the predictor variables. In other words, if we want to calculate the probabilities of \(y = y_1\), it can written as

\begin{align}
P(y = y_1 \mid x_1, x_2, ... , x_n) = \frac{P(y = y_1) \times P(x_1, x_2, ... , x_n \mid y)}{P(x_1, x_2, ... , x_n)} \label{eq:priorprob}
\end{align}

where the left hand side shows \emph{posterior probabilities}, \(P(y = y_1)\) on the right hand side \emph{prior proabibilities} of \(y = y_1\), \(P(x_1, x_2, ... , x_n \mid y)\) \emph{support} data provides for \(y = y_1\), and \(P(x_1, x_2, ... , x_n)\) on the denominator \emph{normalising constant}.

One of the simplest and most popular approaches to the task stated above is \emph{Naive Bayes}\autocite{zhang2004optimality}, where we simply assume that all the predictor variables are indepedent from each other. Then \autoref{eq:priorprob} becomes

\begin{align}
P(y = y_1 \mid x_1, x_2, ... , x_n) = \frac{P(y = y_1) \times \prod_{i=1} P(x_i \mid y)}{P(x_1, x_2, ... , x_n)} \label{eq:naivebayes}
\end{align}

The question is, what if these predictors are actually not completely independent as Naive Bayes assumed? Depending on the subject domain, it might be the case that some of the predictors have dependency relationships. If we assume that all the variables are binary and we need to store all the arbitrary dependencies without any additional assumptions, that means we need to store \(2^n - 1\) values in the memory - which would become too much computational burden even for relatively small number of variables.

This is where Bayesian network comes in - it leverages a graph structure to provide more intuitive representation of conditional dependencies in the domain and allow the user to perform inference tasks in reasonable amount of time and resources.

\begin{definition}
A Bayesian Network \(G = (N,A)\) is a probabilistic graphical model in a directed acyclic graph (DAG) where each node \(n \in N\) represents a variable in the dataset and each arc \((i,j) \in A\) indicates the variable \(j\) being probabilistically dependent on \(i\).\footnote{We are using the neutral expression `probabilistically dependent' as the interpretation of the arcs might become based on the assumptions on the domain. Some researchers interpret \(i\) to be \emph{direct cause} of \(j\), which might not be valid in other cases.}
\label{def:bn_def}
\end{definition}

We can therefore perceive Bayesian network as consisting of two components:
\begin{enumerate}
\item \textbf{\emph{Structure}}, which refers to the directed acyclic graph itself: nodes and arcs that specify dependencies between the variables,
\item \textbf{\emph{Parameters}}, corresponding conditional probabilities of each node (variable) given its parents.
\end{enumerate}

\begin{figure}[h]
\centering
\includegraphics[width=0.6\textwidth]{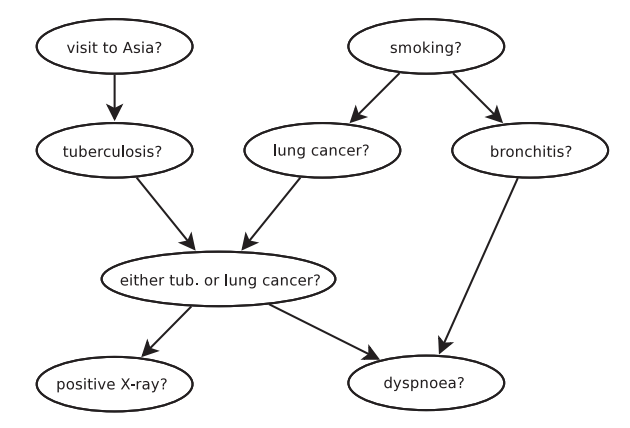}
\caption{\textsf{ASIA} Bayesian Network Structure}
\label{fig:asia_net}
\end{figure}

\noindent One example BN constructed from the \textsf{ASIA} dataset by Lauritzen and Spiegelhalter is presented in \autoref{fig:asia_net}.\autocite{daly2011learning}

\subsection{Key Characteristics of Bayesian Network}
\label{sec:key_characteristics}

\subsubsection{Markov Condition}
The key assumption behind Bayesian network is that one node (variable) is conditionally independent on its non-descendants, \emph{given its parent nodes}. This assumption significantly reduces the space required for inference tasks, since one would need to have only the values of the parents.

To be precise, let's say our Bayesian network tell us that \(x_2\) is a parent node of \(x_1\), but not others. That means \(x_1\) is independent of the rest of the variables \emph{given} \(x_2\). Then we can change \autoref{eq:priorprob} into 

\begin{align}
P(y = y_1 \mid x_1, x_2, ... , x_n) = \frac{P(y = y_1) \times P(x_1 | x_2, y) \times P(x_2, ... , x_n \mid y)}{P(x_1, x_2, ... , x_n)} \label{eq:BNprob}
\end{align}

If we can find other conditional independencies like this using Bayesian network, we can keep changing \autoref{eq:BNprob} above into hopefully smaller number of terms than what we would have had by assuming arbitrary dependencies and independencies.

The assumption explained above is called Markov condition, which the formal definition\autocite{daly2011learning} is stated below:

\begin{definition}
Given a graph \(G = (N,A)\) and a joint probability distribution \(P\) defined over variables represented by the nodes \(n \in N\). If the following statement is also true

\begin{align*}
\forall n \in N, n\independent{}ND(v) \mid Pa(v)
\end{align*}

where \(ND(v)\) refers to the non-descendants and \(Pa(v)\) parent nodes of \(v\), then we can say that G satisfies the Markov condition with P, and (G,P) is a Bayesian network.
\label{def:markov_condition}
\end{definition}

\subsubsection{d-separation}
As stated in \autoref{def:bn_def}, the structure of Bayesian network is a directed acyclic graph: main motivation for using it was to make use of conditional independence to store uncertain information efficiently by associating dependence with connectedness and independence with un-connectedness in graphs. By exploiting the paths on DAG, Judea Pearl introduced a graphical test called \emph{d-separation} in 1988 that can discover all the conditional independencies that are \emph{implied} from the structure (or equivalently Markov condition stated in \autoref{def:markov_condition}):

\begin{definition}
\label{def:trail}
A trail in a directed acyclic graph \(G\) is an undirected path in \(G\), which is a connected sequence of edges in \(G'\) where all the directed edges in \(G\) are replaced with undirected edges.
\end{definition}

\begin{definition}
\label{def:head_to_head}
A head-to-head node with respect to trail \(t\) is a node \(x\) in \(t\) where there are consecutive edges \((\alpha, x)\), \((x, \beta)\) for some nodes \(\alpha\) and \(\beta\) in \(t\). 
\end{definition}

\begin{definition}
(Pearl 1988) If J, K, L are three disjoint subsets of nodes in a DAG \(D\), then \(L\) is said to d-separate \(J\) from \(K\), denoted \(I(J, L, K)_D\), iff there is no trail \(t\) between a node in \(J\) and a node in \(K\) along which (1) every head-to-head node (w.r.t. t) either is or has a descent in \(L\) and (2) every node that delivers an arrow along \(t\) is outside \(L\). A trail satisfying the two conditions above is said to be active, otherwise it is said to be blocked (by \(L\)).\autocite{geiger2013d}
\label{def:d_separation_def}
\end{definition}

Expanding \autoref{def:d_separation_def}, we can summarise the types of dependency relationships extractable from Bayesian network into following\autocite{charniak1991bayesian}\autocite{pearlTears}\autocite{cussenstutorial}:

\begin{figure}[h]
\centering
\includegraphics[width=0.8\textwidth]{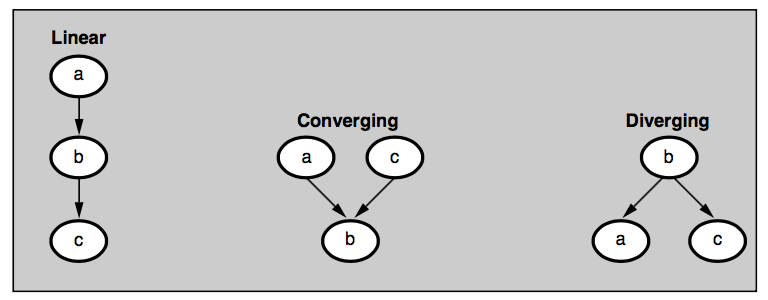}
\caption{Different Connection Types in DAG}
\label{fig:d_separation}
\end{figure}

\begin{itemize}
\item \textbf{Indirect cause}: Consider the connection type `Linear' in \autoref{fig:d_separation}. We can say that \(a\) is independent of \(c\) given \(b\). Although \(a\) might explain \(c\) up to some degree, it becomes conditionally independent when we have \(b\). In light of d-separation definition, we can say that \(a\) \(c\) is d-connected and active when we condition against something other than \(a, b, c\). However, when we consider \(b\) as a condition (member of \(L\) in the definition), \(b\) d-separates \(a\) and \(c\).
\item \textbf{Common effect}: For the `Converging' connection type, we can say that \(a\) and \(c\) becomes dependent given \(b\). \(a\) and \(c\) are independent without \(b\). However, once we condition on \(b\), the two become dependent - once we know \(b\) has occurred, either one of \(a\) and \(c\) would explains away the other since \(b\) is probabilistically dependent on \(a\) and \(c\).
\item \textbf{Common cause}: For the `Diverging' connection type, \(a\) is independent of \(c\) given \(b\).
\end{itemize}

\subsubsection{Markov Equivalent Structures}
Another charcteristic that arise from the structure of Bayesian network is that it might be possible to find another DAG that encodes same set of conditional independencies as original. Two directed acyclic graphs are considered Markov equivalent\autocite{verma1990equivalence} iff they have 

\begin{itemize}
\item same \textbf{skeletons}, which is a graph which all the directed edges of the DAG are replaced with undirected ones, and
\item same \textbf{v-structures}, which refers to all the head-to-head meetings of directed edges without unjoined tails in the DAG.
\end{itemize}

\noindent To represent this equivalence class, we use Partially Directed Acyclic Graph (PDAG) with every directed edge compelled and undirected edge reversible.

\newpage

\section{Integer Linear Programming}

\subsection{Overview}
Integer Linear Programming (ILP) refers to the special group of Linear Programming (LP) problems where the domains of some or all the variables are confined to integers, instead of real numbers in LP. Such problems can be written in the form

\begin{align}
\text{max } &c^\top x \nonumber\\
\text{subject to } &Ax \le b \label{eq:ILP}\\
&x \in \mathbb{Z} \nonumber
\end{align}

If the problem have both non-integer and integer variables, we call it a \emph{Mixed} Integer Linear Programming (MILP) problem. While ILP obviously can be used in situations where only integral quantities make sense, for example, number of cars or people, more effective uses of integer programming often incorporate binary variables to represent logical conditions like \emph{yes or no} decisions.\autocite{williams1999model}

The major issue involved with solving ILPs is that they are usually harder to solve than non-integer LP problems. Unlike conventional LPs which Simplex algorithm can directly produce the solution by checking extreme points of feasible region,\footnote{It is therefore possible that such solution happens to be integers; however, it is generally known from past experience that such case rarely appears in practice.} ILP need additional steps to find a specific solution that are strictly integers. There are mainly two approaches to solving ILP - 1) Branch-and-Bound and 2) Cutting Planes. In fact, these two are not mutually exclusive to be precise but rather can be combined and used together to solve ILP, the approach called 3) Branch-and-Cut. The following subsections will briefly describe the first two and outline the details of the third, which is actually used in the main part of this dissertation.

\subsection{Branch-and-Bound}
The term ``Branch-and-Bound" (BnB) itself refers to an algorithm design paradigm that was first introduced to solve discrete programming problems, developed by Alisa H. Land and Alison G. Doig of the London School of Economics in 1960.\autocite{land1960automatic} BnB have been the basis of solving a variety of discrete and combinatorial optimisation problems,\autocite{clausen1999branch} most notably integer programming.

\begin{figure}[h]
\centering
\includegraphics[width=0.7\textwidth]{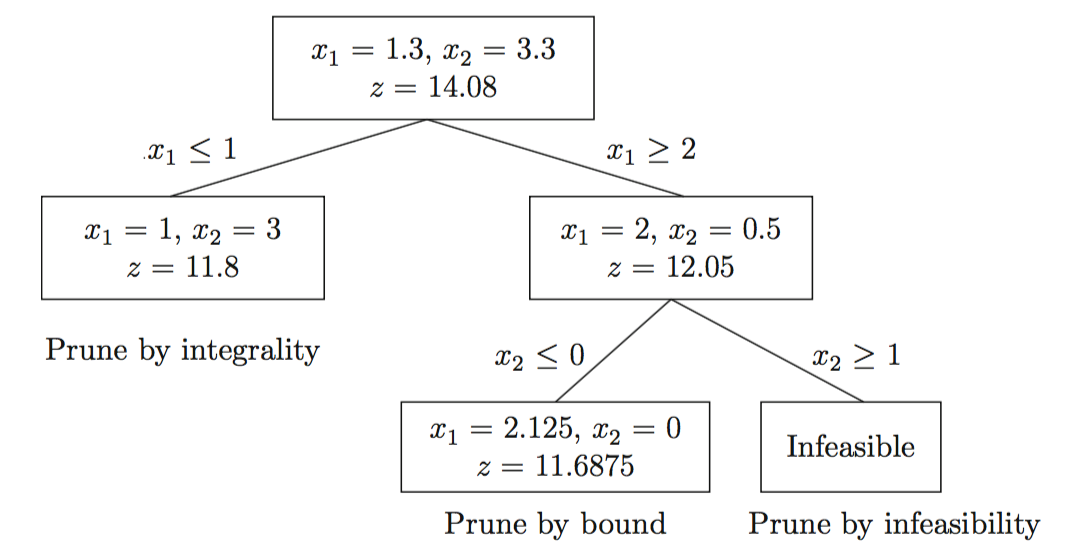}
\caption{Branch-and-Bound example}
\label{fig:bnb_tree}
\end{figure}

BnB in IP can be best described as a divide-and-conquer approach to systematically explore feasible regions of the problem. Graphical representation of this process is shown in \autoref{fig:bnb_tree}.\autocite{conforti2014integer} We start from the solution of LP-relaxed version of the original problem, and choose one of the variables with non-integer solutions - let's call this with variable \(x_i\) with non-integer solution \(f\). Then we create two additional sub-problems (branching) by having

\begin{itemize}
\item one of them with additional constraint that \(x_i \leq \lfloor{}f\rfloor{}\), 
\item another with \(x_i \geq \lceil{}f\rceil{}\).
\end{itemize}

We choose and solve one of the two problems. If the solutions still aren't integers, branch on that as above and solve one of the new nodes (problems), and so on.

\begin{itemize}
\item If one of the new problems return integer solutions, we don't have to branch any more on that node (pruned by integrality), and this solution is called \emph{incumbent} solution - the best one yet.
\item If a new problem is infeasible, we don't have to branch on that as well (pruned by infeasibility).
\item If the problem returns integer solutions but the objective value is smaller than the incumbent, then we stop branching on that node (pruned by bound).
\end{itemize}

When branching terminates on certain node, we go back to the direct parent node and start exploring the branches that haven't been explored yet. After we keep following these rules and there are no more branches to explore, than the incumbent solution at that moment is the optimal solution.

\subsection{Cutting Planes}
Cutting planes is another approach for solving ILP that was developed before BnB. The basic idea is that we generate constraints (which would be a hyperplane) that can cut out the currunt non-integer solution and tighten the feasible region. If we solve the problem again with the new constraint and get integer solutions, then the process ends. If not, we continue adding cutting planes and solve the problem until we get integer solutions.

The question is how we generate the one that can cut out the region as much as possible. There are a number of different strategies available for this task, but the most representative and the one implemented for this project is called \emph{Gomory Fractional Cut}:

\begin{align}
\displaystyle\sum_{j=1}^{n} (a_j - \lfloor{}a_j\rfloor{}) x_j >= a_0 - \lfloor{}a_o\rfloor{} \label{eq:gomory_cut_equation}
\end{align}

\noindent The equation above is called Gomory fractional cut,\autocite{gomory1963algorithm}\autocite{conforti2014integer} where \(a_j, x_j\) comes from the row of the optimal tableau.  Since \(\sum_{j=1}^{n} a_j x_j = a_0\), there must be a \(k \in \mathbb{Z}\) that satisfies \(\sum_{j=1}^{n} (a_j - \lfloor{}a_j\rfloor{}) x_j = a_0 - \lfloor{}a_o\rfloor{} + k\). Also, \(k\) is non-negative since \(\sum_{j=1}^{n} (a_j - \lfloor{}a_j\rfloor{}) x_j\) is non-negative. Therefore \autoref{eq:gomory_cut_equation} holds. As a result, Gomory factional cut can cut out the current fractional solution from the feasible region.

In additional to general purpose cuts like Gomory cuts that can be applied to all the IP problems, we might be able to generate cutting planes that are specific to the problem. This dissertation also examines some domain-specific cutting planes for the BN structure learning problem, which is done in \autoref{ch:solutions}.

\subsection{Branch-and-Cut}
Branch-and-Cut combines branch-and-bound and cutting planes into one algorithm, and it is the most successful way of solving ILP problems to date. Essentially following the structure of BnB to explore the solution space, we add cutting planes whenever possible on LP-relaxed problems before branching to tighten the upper bound, so that we can hopefully branch less than the standard BnB. The decision of whether to add cutting planes depends on the specific problems and success of previously added cuts. Pseudocode of Branch-and-Cut algorithm is presented in \autoref{alg:branch_and_cut}. The criteria used for our problem will also be discussed in \autoref{ch:solutions}.

\begin{algorithm}[H]                      
\caption{Branch-and-Cut Algorithm}     
\label{alg:branch_and_cut}             
\begin{algorithmic}                    
    \REQUIRE initial\_problem, problem\_list, objective\_upper\_bound, best\_solution
    \ENSURE best\_solution \(\in \mathbb{Z}\)
    \STATE \(\text{objective\_lower\_bound} \Leftarrow - \infty\)
    \STATE \(\text{problem\_list} \Leftarrow \text{problem\_list} \cup \text{initial\_problem}\)
    \WHILE{problem\_list \(\not= \emptyset\)}
        \STATE current\_problem \(\Leftarrow p \in \text{problem\_list}\)
        \STATE Solve LP-Relaxed current\_problem
        \IF{infeasible}
            \STATE Go back to the beginning of the loop
        \ELSE
            \STATE \(z \Leftarrow \text{current objective}\), \(x \Leftarrow \text{current solution}\)
        \ENDIF
        \IF{\(z \leq \text{objective\_upper\_bound}\)}
            \STATE Go back to the beginning of the loop
        \ENDIF
        \IF{\(x \in \mathbb{Z}\)}
            \STATE best\_solution \(\Leftarrow x\)
            \STATE objective\_upper\_bound \(\Leftarrow z\)
            \STATE Go back to the beginning of the loop
        \ENDIF
        \IF{cutting planes applicable}
            \STATE Find cutting planes violated by \(x\)
            \STATE add the cutting plane and go back to `Solve LP-Relaxed current\_problem'
        \ELSE
            \STATE Branch using non-integer solutions in \(x\)
            \STATE \(\text{problem\_list} \Leftarrow \text{problem\_list} \cup \text{new\_problems}\)
        \ENDIF
    \ENDWHILE
    \RETURN best\_solution
\end{algorithmic}
\end{algorithm}

\chapter{Learning the Structure}
\label{ch:problem_intro}

In some cases, it might be possible to create a Bayesian network manually if there's a subject expert who already have a knowledge about the relationships between the variables. Many statistical softwares such as OpenBUGS\footnote{\url{http://www.openbugs.net}} and Netica\footnote{\url{https://www.norsys.com/netica.html}} allow the users to specify BN models.

However, it would be impossible to create such models in other cases, either because we simply do not have such domain knowledge or there are too many variables to consider. Therefore, there have been constant interests in ways to automatically learn the Bayesian network structure that best explains the data.

Chickering (1996) have showed that the structure learning problem of Bayesian network is NP-hard.\autocite{chickering1996learning} Even with additional conditions such as having an independence oracle or the number of parents limited to 2, the problem still remains intractable.\autocite{chickering2004large}\autocite{dasgupta1999learning} Therefore, the efforts have been focused on making the computation feasible using various assumptions and constraints while attempting to provide some guarantee of optimality. 

Before going into the actual solutions, the first thing we need to consider is how we should define the best structure. Since Bayesian network is about revealing conditional independencies in the domain, the best Bayesian network should be able to identify as many dependencies as possible that are highly likely in terms of probability.

Researchers formalised this notion as \emph{score-and-search}, where we search through the space of structure candidates with scores for each of them, and select the highest scoring one. Section \ref{sec:score_metric} explains how such score metric works. After explaining the score metric, our integer linear programming formulation of finding the best Bayesian network structure is presented in section \ref{sec:ilp_formulation}.

\section{Score Metrics}
\label{sec:score_metric}
By scoring a candidate BN structure, we are measuring a probability of the candidate being the BN structure representing joint probability distribution (JPD) that our training data is sampled from. Following Bayes' Theorem, we want to calculate the \emph{posterior} probability

\begin{align}
\label{eq:posterior_structure}
P(B_S^h \mid D) = c \times P(B_S^h) \times P(D \mid B_S^h)
\end{align}

where \(B_S^h\) is a hypothesis that the candidate structure \(B_s\) represents the JPD that the dataset \(D\) was sampled from, \(P(B_S^h)\) being the \emph{prior} probability of the hypothesis, \(P(D \mid B_S^h)\) \emph{support} \(D\) provides for the hypothesis, and \(c\) is a normalising constant.

For the priors of each candidate hypothesis, a uniform prior is usually assumed, even though it might be possible to assign different priors based on expert knowledge.\autocite{heckerman1995likelihoods} The main problem here is how to measure the support \(P(D \mid B_S^h)\) - more generally, how we can measure the goodness of fit of the candidate structure to the dataset. There are a number of score metrics based on different theoretical foundations, but this dissertation uses BDeu metric, the special type of Bayesian Dirichlet-based metrics which their inner workings are explained below:

\subsection{Bayesian Dirichlet}
Bayesian Dirichlet (BD)-based score metrics are Bayesian approaches to scoring using Dirichlet distribution to calculate the posterior probability of the structure. These score metrics have the following assumptions in common\autocite{carvalho2009scoring}:

\subsubsection{Notation}
\begin{itemize}
\item \(\Theta_G = \{\Theta_{i}\}\) for \(i = 1, ... , n\): Set of parameters of all the variables \(i = 1, ... , n\) in Bayesian Network DAG \(G\)
\item \(\Theta_i = \{\Theta_{ij}\}\) for \(j = 1, ... , q_i\): Set of parameters of all the \(q_i\) parent configurations \(j = 1, ... , n\) for just one variable \(i\)
\item \(\Theta_{ij} = \{\theta_{ijk}\}\) for \(k = 1, ... , r_i\): Set of parameters (physical probabilities) of \(i\) taking each \(r_i\) number of values, given one parent configuration \(j\)
\end{itemize}

\subsubsection{Assumptions}
\begin{enumerate}
\item \textbf{Multinomial Samples}: Dataset D have multinomial samples with yet unknown physical probabilities \(\theta_ijk\).
\item \textbf{Dirichlet Distribution}: Set of physical probabilities in \(\Theta_{ij} = \theta_{ij1}, \theta_{ij2}, ... , \theta_{ijr_i}\) follows Dirichlet distribution.
\item \textbf{Parameter Independence}: All the \(\Theta_i\) are independent with each other (global parameter independence), and all the \(\Theta_{ij}\) are independent with each other as well (local parameter independence).
\item \textbf{Parameter Modularity}: Given two BN structures \(G\) and \(G'\), if \(i\) have same set of parents in both structures, then \(G\) and \(G\) have identical \(\Theta_{ij}\).
\end{enumerate}

Before going deeper, let's briefly go over the rationale behind using a Dirichlet distribution. If we were certain about the values of all the \(\theta_{ijk}\), we could simply just express them as a multinomial distribution. However, since we don't have such information, we estimate the behaviour of such distribution by using Dirichlet distribution, which allows us to reflect our beliefs about each \(\theta_{ijk}\) using corresponding parameters \(\alpha_{ijk}\), which are constructed \emph{before} taking our data \(D\) into account.

While it is theoretically possible to use a distribution other than Dirichlet, we use them as it is algebraically straightforward to calculate posterior probabilities. First, since \(\Theta_{ij}\) follows Dirichlet distribution, we know

\begin{align}
\label{eq:dirichlet_general}
p(\Theta_{ij} \mid B_S^h, \alpha) = \frac{\Gamma(\sum_{k=1}^{r_i} \alpha_{ijk})}{\prod_{k=1}^{r_i}} \times \prod_{k=1}^{r_i} \theta_{x=k}^{\alpha_{ijk} - 1}
\end{align}

Also, expected value \(E(\theta_{ijk})\) is \(\frac{\alpha_{ijk}}{\alpha_0}\), where \(\alpha_0 = \sum_{ijk} \alpha_{ijk}\). 

With multinomial sample \(D\), We want to calculate

\begin{align}
\label{eq:dirichlet_givenD}
p(\Theta_{ij} \mid D, B_S^h, \alpha) = c \times \prod_{k=1}^{r_i} p(\Theta_{ij} \mid B_S^h, \alpha) \times \theta_{ijk}^{N_{ijk}}
\end{align}

Since we have \autoref{eq:dirichlet_general}, we can rewrite \autoref{eq:dirichlet_givenD} as

\begin{align}
\label{eq:dirichlet_givenD_conjugate}
p(\Theta_{ij} \mid D, B_S^h, \alpha) = c \times \prod_{k=1}^{r_i} \theta_{ijk}^{(\alpha_{ijk}+ N_{ijk}) - 1}
\end{align}

where \(N_{ijk}\) is a number of times \(ijk\) appeared in \(D\).

So \autoref{eq:dirichlet_givenD_conjugate} shows that the posterior distribution of \(\Theta_{ij}\) given \(D\) is also a Dirichlet distribution. We say Dirichlet distribution is a \emph{conjugate prior} to multinomial samples, where both prior and posterior distributions are Dirichlet. Also, we now have

\begin{align}
E(\theta_{ijk} \mid D, B_S^h, \alpha) = \frac{\alpha_{ijk} + N_{ijk}}{\alpha_{ij} + N_{ij}} \text{, where } &N_{ij} = \sum_{ijk} N_{ijk} \label{eq:expected_ijk_given_D}
\end{align}

\subsubsection{BD metric}

With \autoref{eq:expected_ijk_given_D}, we can also calculate the probability of seeing a certain combination of all the variable values \(C_{m+1}\), which can be written as

\begin{align}
\label{eq:c_m1_prob}
p(C_{m+1} \mid D, B_S^h, \alpha) = \prod_{i=1}^{q} \prod_{j=1}^{q_i} \frac{\alpha_{ijk} + N_{ijk}}{\alpha_{ij} + N_{ij}}
\end{align}

\autoref{eq:c_m1_prob} makes our original task of calculating the support of dataset to candidate hypothesis \(P(D \mid B_S^h, \alpha)\) algebraically convenient. Let's say our dataset \(D\) have \(m\) instances, then

\begin{align}
\label{eq:support_dataset}
P(D \mid B_S^h, \alpha) = \prod_{d=1}^m p(C_d \mid C_1, ... , c_{d-1}, B_S^h, \alpha)
\end{align}

where each \(C_d\) represents an instance in \(D\) with certain combination of variable values. Expanding \ref{eq:c_m1_prob}, \autoref{eq:support_dataset} can be calculated by

\begin{align}
\label{eq:support_dataset_long_form}
P(D \mid B_S^h, \alpha) &= \prod_{i=1}^{n} \prod_{j=1}^{q_i} \bigg\{
\bigg[ \frac{\alpha_{ij1}}{\alpha_{ij}} \times \frac{\alpha_{ij1} + 1}{\alpha_{ij} + 1} \times ... \times \frac{\alpha_{ij1} + (N_{ij1} - 1)}{\alpha_{ij} + (N_{ij1} - 1)} \bigg] \\ \nonumber
&\times \bigg[ \frac{\alpha_{ij2}}{\alpha_{ij} + N_{ij1}} \times \frac{\alpha_{ij2} + 1}{\alpha_{ij} + N_{ij1} + 1} \times ... \times \frac{\alpha_{ij2} + (N_{ij2} - 1)}{\alpha_{ij} + N_{ij1} + (N_{ij2} - 1)} \bigg] \\ \nonumber
&\times \bigg[ \frac{\alpha_{ijr_i}}{\alpha_{ij} + \sum_{k=1}^{r_i - 1} N_{ijk}} \times \frac{\alpha_{ijr_i} + 1}{\alpha_{ij} +  \sum_{k=1}^{r_i - 1} N_{ijk} + 1} \times ... \times \frac{\alpha_{ijr_i} + (N_{ijr_i} - 1)}{\alpha_{ij} + (N_{ij} - 1)} \bigg] \bigg\} \\ \nonumber
&= \prod_{i=1}^{n} \prod_{j=1}^{q_i} \frac{\Gamma(\alpha_{ij})}{\Gamma(\alpha_{ij} + N_{ij})} \times \prod_{k=1}^{r_i} \frac{\Gamma(\alpha_{ijk} + N_{ijk})}{\Gamma(\alpha_{ijk})}
\end{align}

Therefore, the probability of the hypothesis \(P(D \mid B_S^h, \alpha)\) can be calculcated using \ref{eq:support_dataset_long_form}, and if we log-transform it we get

\begin{align}
\label{eq:bd_score}
BD(B, D) = log(P(B_S^h \mid \alpha)) + \sum_{i=1}^{n} \sum_{j=1}^{q_i} \bigg(log \bigg(\frac{\Gamma(\alpha_{ij})}{\Gamma(\alpha_{ij} + N_{ij})}\bigg) + \sum_{k=1}^{r_i} log \bigg(\frac{\Gamma(\alpha_{ijk} + N_{ijk})}{\Gamma(\alpha_{ijk})} \bigg) \bigg)
\end{align}

\autoref{eq:bd_score} is called \emph{BD scoring function}, introduced by Heckerman, Geiger and Chickering.\autocite{heckerman1995learning}

\subsubsection{K2, BDe and BDeu}
While BD metric is logically sound, it is practically unusable as we need to have all the \(\alpha_{ijk}\) in hand to calculate the total score. K2 metric in \autoref{eq:k2_score} by Cooper and Herskovits (1992)\autocite{cooper1992bayesian}, which was actually developed before BD metric, simply assigns \(\alpha_{ijk} = 1\) for all \(ijk\), assuming uniform Dirichlet distribution (not uniform distribution) on the prior.

\begin{align}
\label{eq:k2_score}
K2(B, D) = log(P(B)) + \sum_{i=1}^n \sum_{j=1}^{q_i} \bigg( log \bigg( \frac{(r_i - 1)!}{(N_{ij} + r_i - 1)!} \bigg) + \sum_{k=1}^{r_i} log(N_{ijk}!) \bigg)
\end{align}

BDe metric\autocite{heckerman1995learning} attempts to reduce the number of \(\alpha_{ijk}\) that needs to be specified by introducing likelihood equivalence. Let's suppose there's a complete BN structure \(G\) that specifies all the `true' conditional dependencies without any missing edges, and our candidate structure \(B_S^h\) is Markov equivalent to it. Then their parameters for Dirichlet distribution and hence their likelihood should be the same as they represent same joint probability distribution. Then we can get \(\alpha_{ijk} = \alpha' \times P(X_i = x_{ik}, \prod_{X_i} = w_{ij} \mid G)\), where \(\alpha'\) represent the level of belief on the prior and \(w_{ij}\) is j-th configuration of parents for \(i\) in \(G\). In other words, we only need \(\alpha_{ijk}\) for the configuration that are actually represented by edges in \(B_S\).

BDeu metric\autocite{Buntine1991} goes one step further from BDe by simply assuming that \(P(X_i = x_{ik}, \prod_{X_i} = w_{ij} \mid G) = \frac{1}{r_i \times q_i}\), which assigns uniform probabilities to all combinations of \(x_i\) values and its parent configurations. This allows the user to calculate the scores with limited prior knowledge, while maintaining the property of likelihood equivalence which K2 metric cannot.

\subsubsection{Decomposability of Score Metrics}
As seen from \autoref{eq:bd_score}, these score metrics are log-transformed to provide decomposability. Since the outermost sum is summed over each variable \(i\), we can say that BDeu score is a sum of \emph{local scores} for each variable given their parent node configurations. Therefore, we can say that we are trying to find the best structure with the highest score by choosing the parent nodes for each variable that maximises their local scores. This goal is expressed as the objective in \autoref{eq:cussens_ilp_objective}.


\newpage

\section{ILP Formulation}
\label{sec:ilp_formulation}
Based on the decomposable score metrics in \autoref{sec:score_metric}, we present integer linear programming formulation of finding the best Bayesian network structure. One of the early works on the ILP formulation have been done by Cussens\autocite{Cussens} for reconstructing pedigrees (`family tree') as a special type of BN.

The work independently done by Jaakkola et al.\autocite{jaakkola2010learning} provides proof that the well known acylic subgraph polytope \(P_{dag}\) is not tight enough for the BN structure learning problem and presents tighter constraint of maintaining acyclicity dubbed cluster constraint, along with the algorithm to approximate the tightened polytope using duals.

Cussens incorporated the findings of \autocite{jaakkola2010learning} into branch-and-cut algorithm by adding cluster constraints as cutting planes, along with general purpose cutting planes and heuristics algorithm for speeding up the performance.\autocite{Cussens2012}\autocite{bartlett2015integer}. This dissertation largely follows the work of Barlett and Cussens\autocite{bartlett2015integer}, but with slightly different strategies for finding cutting planes and object-oriented implementation of experiment program from scratch using \textsf{Python} language and \textsf{Gurobi} solver, which is completely independent from Cussens's computer program \textsf{GOBNILP} written in \textsf{C} language\autocite{Cussens2015}\footnote{\url{https://www.cs.york.ac.uk/aig/sw/gobnilp/}}.

\subsection{Sets, Parameters, Variables}
\begin{itemize}
\item \(v = \text{node in BN}\)
\item \(W = \text{parent set candidate for } v\)
\item \(c(v, W) = \text{local scores for } v \text{ having } W \text{ as parents}\)
\item \(I(W \rightarrow v) = \text{binary variable for } W \text{ being selected for } v\).
\end{itemize}

\subsection{Objective}
\begin{align}
\label{eq:cussens_ilp_objective}
\text{\textbf{maximise} } \text{total score: } \sum_{v, W} c(v, W) \times I(W \rightarrow v)
\end{align}

\subsection{Constraints}

\subsubsection{Only One Parent Set Constraint}
\begin{align}
\label{eq:cussens_only_one_parent}
\forall v \in V: \sum_{W} I(W \rightarrow v) = 1
\end{align}

\subsubsection{Acyclicity Constraint}
Since Bayesian network takes a form of DAG, the constraint to ensure acyclicity is the primary challenge of this ILP formulation. Acyclicity constraints on directed acyclic graphs for use in linear programming have been studied quite extensively as facets of acylic subgraph polytope \(P_{dag}\). Typically, such constraints can be expressed in the form

\begin{align}
\label{eq:cycle_cuts}
\sum_{(i,j) \in C} x_{ij} \leq |C| - 1
\end{align}

\noindent where \(x_{ij}\) is a binary variable for each directed edge \((i,j)\) and \(C\) is any subset of nodes.

While this constraint is obviously valid for Bayesian network structures as well and actually used as cutting planes in \autoref{sec:cycle_cutting_planes}, this is not tight enough for our model for two reasons. First, it is already known that \ref{eq:cycle_cuts} still allows edge selections that are actually not valid except for some special cases of planar DAG\autocite{grotschel1985acyclic}.

More importantly however, our binary variable selects a group of parent nodes but not individual edges. This creates another problem as simplying having just \ref{eq:cycle_cuts} would allow a situation where the choice of parent nodes for one variable gets divided over two candidate sets.

\begin{proposition}
\label{prop:cycle_cut_not_enough}
There exists some cases of dicycles in Bayesian network G that cannot be cut off by the constraint in \autoref{eq:cycle_cuts}.
\end{proposition}

\begin{proof}
We prove by counterexample. Let's suppose we have variable nodes \(N = {A,B,C}\). If the solver assigned \(0.5\) to the parent set choices \((A \mid B, C)\) \((B \mid A, C)\), \((C \mid A, B)\) and 0.5 to \((A \mid \emptyset)\), \((B \mid \emptyset)\), \((C \mid \emptyset)\), then these still satisifies \ref{eq:cussens_only_one_parent} by \(0.5 + 0.5 = 1\) and \ref{eq:cycle_cuts} by \(0.5 + 0.5 + 0.5 = 1.5 < 3\) for each node, but this solution does not represent any valid BN structure.\qedhere
\end{proof}

\begin{align}
\label{eq:cussens_acyclicity}
\forall \text{ cluster } C \subseteq V: \sum_{v \in C} \sum_{W \cap C = \emptyset} I(W \rightarrow v) \geq 1
\end{align}

\autoref{eq:cussens_acyclicity} is a constraint developed by Jaakkola et al.\autocite{jaakkola2010learning} to overcome this issue. The basic idea is that for every possible cluster (subset of nodes) in the graph, there should be at least one node that does not have any of its parents in the same cluster, or have no parents at all. This would be able to enforce acyclicity as there would be no edges at all that points to such node within the cluster (but might have nodes that start from this node to the other node in the cluster). Since this constraint applies to clusters of every size in the graph, we can see that this constraint is much tighter than \ref{eq:cycle_cuts}.

Since both \autoref{eq:cycle_cuts} and \autoref{eq:cussens_acyclicity} would require exponential number of constraints if we add every possible cases of them at once into our model, we instead add them as cutting planes if needed. We first solve LP relaxation without them, search for constraints that are violated by the current solution, then add those to the model and solve again. The details of how we search for these cutting planes are explained in \autoref{ch:solutions}.

\chapter{Finding Solutions}
\label{ch:solutions}
Based on the formulation outlined in \autoref{sec:ilp_formulation}, we present the actual process developed to solve the problem. Due to the number of acyclicity constraints that grows exponentially with respect to the number of variables, we do not add them directly to the formulation as that will complicate the shape of our feasible region and make the solution process difficult, let alone the time of generating all the constraint statements. Instead, we add them as cutting planes on demand - we first solve the problem without those constraints and search for only the constraints that are actually violated by the current solution. In addition, we considered some heuristic algorithm that takes advantage of the relaxed solution to make our solving process faster. 

\section{Cluster Cuts}
\label{sec:cluster_cuts_sub_ip}
To add \autoref{eq:cussens_acyclicity} as cutting planes, we need to find a cluster or set of nodes that violates \ref{eq:cussens_acyclicity} under the solution yet without this constraint. Since cutting planes allow us to tighten our feasible region and speed up the solution process, we want to look for the cluster that violates the constraint the most, i.e. the one that has the most members with their parents within the same cluster. To express this idea formally, let's first see that \autoref{eq:cussens_acyclicity} can be rewritten as

\begin{align}
\label{eq:cussens_acyclicity_reexpressed}
\forall C \subseteq V: \sum_{v \in C} \sum_{W: | W \cap C | \geq 1} I(W \rightarrow v) \leq |C| - 1
\end{align}

\noindent This is because we have \autoref{eq:cussens_only_one_parent}, where we should have exactly one parent set choices for each node. There should be exactly \(|C|\) parent set choices, but at least one of them should choose a parent set that is \emph{completely} outside \(C\). Therefore, The number of \(I(W \rightarrow v)\) that chooses \(W\) with one or more of its members inside \(C\) (\(| W \cap C | \geq 1\)) should be limited to \(|C| - 1\).

So we are looking for a cluster \(C\) that makes the LHS of \ref{eq:cussens_acyclicity_reexpressed} exceed the RHS the most. Cussens\autocite{Cussens2015} have suggested one way of achieving this by formulating this as another small IP problem based on the current relaxed solution. For each non-zero \(I(W \rightarrow v)\) in current relaxed solution, we create corresponding binary variable \(J(W \rightarrow v)\). We also put these non-zero solutions as coefficients of \(J(W \rightarrow v)\), denoted as \(x(W \rightarrow v)\) below. Lastly, we create binary variables \(M(v \in C)\) for all the nodes \(v\), which will indicate that each node is chosen to be included.

\begin{align}
\text{\textbf{Objective: } } &\text{maximise } \sum_{v, W} x(W \rightarrow v) \times J(W \rightarrow v) - \sum_{v \in V} M(v \in C) \\
\text{\textbf{Constraint: }}& \nonumber \\ 
&M(v \in C) = 1 \text{ for each } J(W \rightarrow v) = 1 \\
&M(w \in C) = 1 \text{ for at least one } w \in W \text{ for each } J(W \rightarrow v) = 1
\label{eq:cussens_cluster_sub_ip_formulation}
\end{align}

\noindent We are having \(x(W \rightarrow v)\) as coefficients here because we want to choose the cluster that is supported by the current relaxed solution. That is, we want the choice of \(I(W \rightarrow v)\) to be the one that cuts out the feasible region the most including the current relaxed solution, not just some arbitrary space in the feasible region. We also have \(\sum_{v \in V} M(v \in C)\) that represents the size of the found cluster \(C\). Since we are trying to get a single cluster that have more \(I(W \rightarrow v) > 0\) than the total number of nodes, we want \(\sum x(W \rightarrow v) \times J(W \rightarrow v)\) to exceed \(\sum_{v \in V} M(v \in C)\) as much as possible. In addition, we are ruling out any solution with the objective value \(\leq -1\) to avoid any unviolated clusters.

The two constraints are the representation of \(I(W \rightarrow v)\) as node in the cluster. For any cluster with \(I(W \rightarrow v)\), then \(v\) must be inside such cluster. For members of \(W\), at least one of them should be in the same cluster. \autocite{Cussens2015} have implemented these two constraints by using their SCIP solver's \textsf{logicor} functionality, which is based on constraint programming. In more generic linear programming convention, we rather express these by

\begin{align}
\forall J(W \rightarrow v): J(W \rightarrow v) - M(v \in C) = 0 \\
\forall J(W \rightarrow v): J(W \rightarrow v) \leq \sum_{w \in W} M(w \in C)
\end{align}

\noindent where the first constraint simply says \(M(v \in C)\) must be 1 if \(J(W \rightarrow v) = 1\), and the second constraint forces at least one \(M(w \in C)\) to be 1 if \(J(W \rightarrow v) = 1\).

We pass this formulation to the solver to get the best cluster cuts possible. When the solver returns the solution with the values of each \(M(v \in C)\), we now have a cluster \(C\) to add to the main model. We add additional constraint as expressed in \autoref{eq:cussens_acyclicity_reexpressed}, but only for the particular cluster found by the above model.

Given that this is a relatively simple ILP problem, we can obtain the cluster cut in a very short amount of time in conjunction with a fast solver. While there is an alternative approach that formulates this problem as an all-pairs shortest path algorithm\autocite{jaakkola2010learning} based on the same ideas explained here, we chose to use ILP solvers as it was more practical to achieve faster solving process with the solver than trying to solve this problem with an algorithm written in Python.

\section{Cycle Cuts}
\label{sec:cycle_cutting_planes}
We add exactly one cluster cuts each time we solve the sub-IP problem. However, it might be also helpful to rule out all the cycles that can be directly detected in the current solution and tighten the feasible region. That is, we want to add \autoref{eq:cycle_cuts} as cutting planes as well. In fact, all the cycles can be directly converted as cluster cuts, since all the clusters simply refer to any set of nodes with a limit on the number of edges. However, the converse is not true since there might be cluster cuts that might not be shown as cycles, as proved in \autoref{prop:cycle_cut_not_enough}.

In order to add cycle cuts, we first need to find all the existing cycles in the current solution. There are several different approaches to acheive this, but we used the algorithm developed by Johnson\autocite{johnson1975finding}, which is currently one of the best known universally applicable version. After we get all the unique elementary cycles, we go over each of them to get \autoref{eq:cussens_acyclicity_reexpressed} over the members of each cycle.

\section{Sink Finding Heuristic}
With cluster cuts and cycle cuts, we now can obtain optimal bayesian networks by solving the ILP model. However, depending on the size of dataset and the number of variables, it might be too time consuming to wait until the program reaches the optimal solution. Rather, we might want to get sub-optimal but feasible solutions that might be reasonably close to the optimal one. This also allows us to get a better lower bound on the problem and prevent excessive branchings on the solver.

\begin{figure}[h]
\centering
\includegraphics[width=0.85\textwidth]{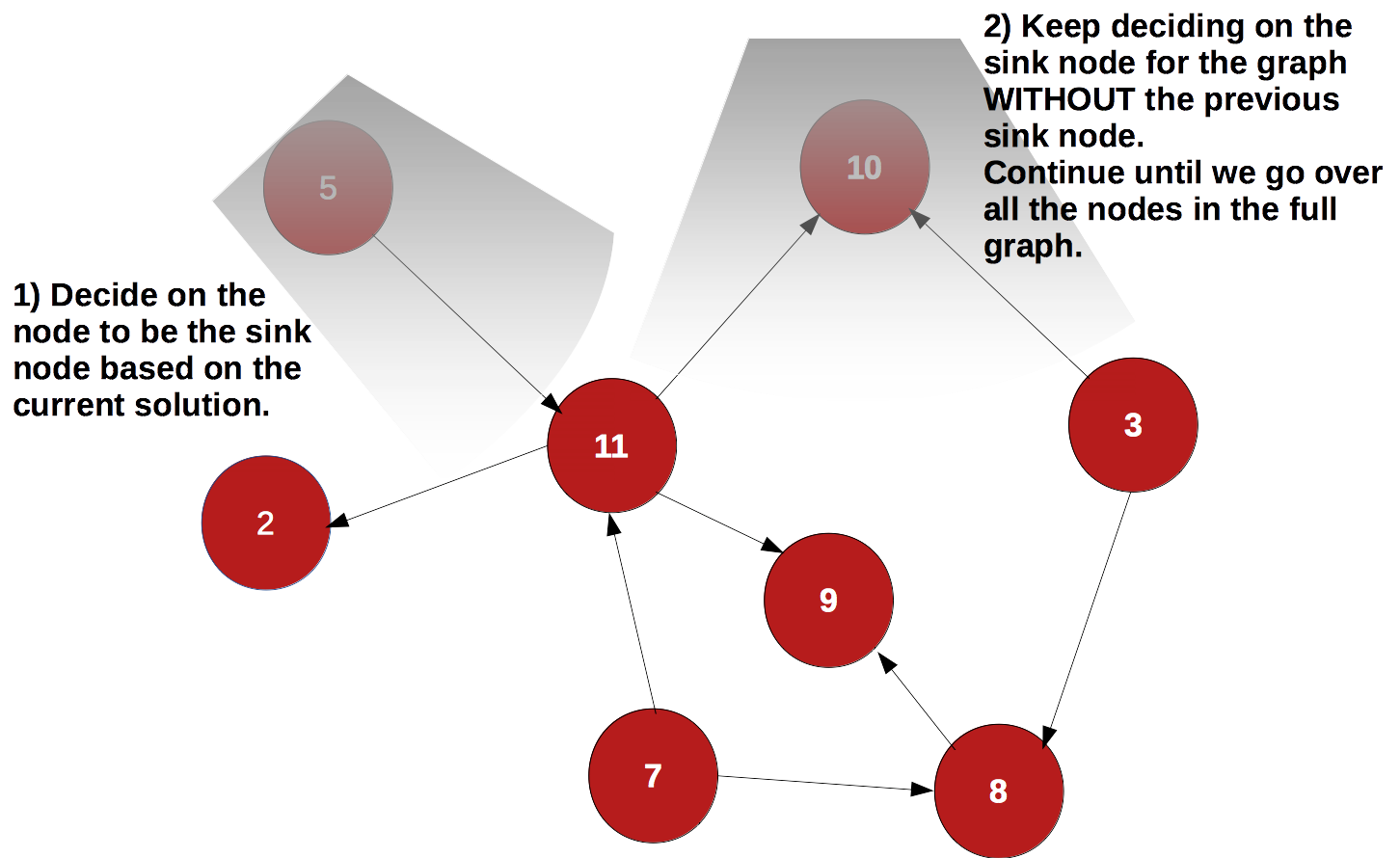}
\caption{Graphical Illustration of Using Sink Nodes to Construct a DAG.}
\label{fig:sink_finding_graphics}
\end{figure}

\autocite{Cussens2015} have suggested an idea for a heuristic algorithm to acquire such solution, which is based on the fact that every DAG has at least one sink node, or the node that has no outgoing edge. Such sink node in a Bayesian network structure would indicate that the variable would not directly influence others, and we can freely choose a parent set for that variable node without creating any cycles. Let's suppose that we remove that node and its incoming edges. In the resulting DAG, there should be another sink node to maintain acyclicity, which we can again choose a parent set for it. If we keep following this rule and decide on all the variables we have, a feasible DAG is constructed. We illustrate this idea graphically in \autoref{fig:sink_finding_graphics}.

So we are essentially adding nodes one by one to construct a DAG. We want to add them in an order that maximises the total score. During the first time we decide on the node to assign parents, we check the local scores of all the parents and pick the highest scoring set for each node. We then look at the current relaxed solution and rank these node/parent combinations based on their closeness to 1. We assign 1 to the variable representing the combination that are closest to 1. Since this node is a sink node, we must make sure that there's no other nodes where this node becomes a parent. We check all the parent candidates of other nodes and assign 0 to all the variables representing parents with this node.

All the following iterations would be identical as the first except we rule out all the parents that are already assigned 0 in previous iterations when picking the highest scoring parent candidate. These procedures are fully outlined in \autoref{alg:sink_finding}.

\begin{algorithm}                    
\caption{Sink Finding Heuristic Algorithm}     
\label{alg:sink_finding}             
\begin{algorithmic}                  
    \REQUIRE current\_solution
    \ENSURE \(\text{nodes\_to\_decide} \Leftarrow 0\)
    \STATE \(\text{heuristic\_BN} \Leftarrow \text{empty list}\)
    \WHILE {\(\text{nodes\_to\_decide} > 0\)}
        \STATE \(\text{best\_parents} \Leftarrow \text{empty list}\)
        \STATE Get all the local scores for this node sorted in descending order
        \FOR {each sorted parent candidates \(W\) of this node}
            \IF{\(W\) not in heuristic\_BN}
                \STATE \(\text{best\_parent\_for\_this\_node} \Leftarrow W\)
                \STATE break
            \ENDIF
        \ENDFOR
        \STATE \(\text{best\_parents} \cup \text{best\_parent\_for\_this\_node}\)
        \STATE \(\text{best\_distance} \Leftarrow \infty\)
        \FOR {chosen \(W \in \text{best\_parents}\)}
            \STATE \(\text{distance\_of\_W} \Leftarrow (1.0 - \text{current\_solution(W)})\)
            \IF {\(\text{distance\_of\_W} < \text{best\_distance}\)}
                \STATE \(\text{best\_distance} \Leftarrow \text{distance\_of\_W}\)
            \ENDIF
        \ENDFOR
        \STATE \(\text{heuristic\_BN(best\_distance)} \Leftarrow 1\)
        \FOR {all other nodes}
            \STATE \(\text{heuristic\_BN(parent)} \Leftarrow 0\) for all parents with W
        \ENDFOR
        \FOR {all other parents of this node}
            \STATE \(\text{heuristic\_BN(other parents)} \Leftarrow 0\)
        \ENDFOR
    \ENDWHILE
    \RETURN heuristic\_BN
\end{algorithmic}
\end{algorithm}

\chapter{Implementation and Experiments}
Based on the ILP formulation with problem-specific cutting planes and heuristic algorithm developed in \autoref{ch:problem_intro} and \ref{ch:solutions}, we present the details of our computer program implementation written in \textsf{Python} language with \textsf{Gurobi} solver and examine its performance on the reference datasets.

\section{Implementation Details}
\label{sec:implementation_details}
\begin{figure}[h]
\centering
\includegraphics[width=0.9\textwidth]{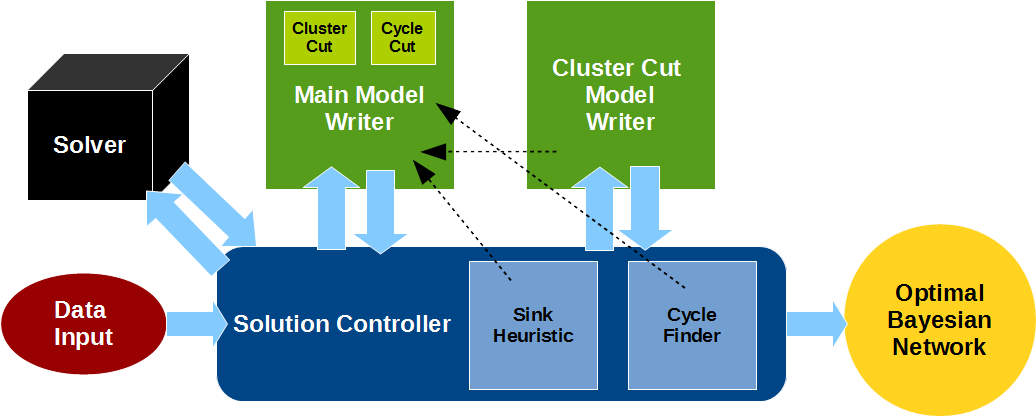}
\caption{\textsf{Bayene} Program Design}
\label{fig:bayene_structure}
\end{figure}

We created a \textsf{Python} package named \textsf{Bayene} \textsf{(\textbf{Baye}sian \textbf{ne}twork)} that discovers an optimal Bayesian network structure based on our ILP formulation given the data input. Refer to \autoref{fig:bayene_structure} for the overall structural design of \textsf{Bayene}. We have \textsf{solution controller} that takes charge of controlling the overall solving process and facilitating communication between the model writer and solver. Two model writers, one for the main model and another for the cluster cut finding model, transforms ILP problems as programming objects that can be transferred to the solver and always modifiable through dedicated functions when needed. Model writers and interfaces to solvers are written with \textsf{Pyomo} package\autocite{hart2011pyomo}. In addition, \textsf{solution controller} includes our own sink-finding heurisitic algorithm, and a cycle detector based on the elementary cycle finding algorithm implementation from \textsf{NetworkX} package\autocite{hagberg-2008-exploring}.

\begin{figure}[h]
\centering
\includegraphics[width=0.9\textwidth]{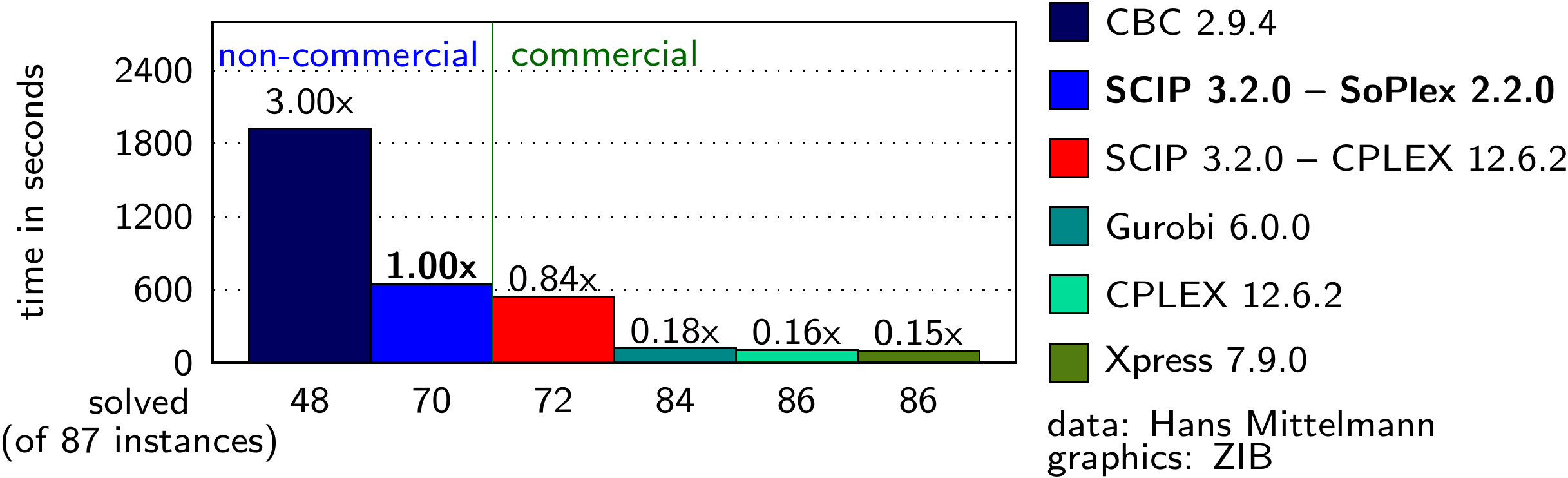}
\caption{Comparison of Different ILP Solvers from \textsf{SCIP} website}
\label{fig:ilp_solvers_comparison}
\end{figure}

\textsf{GOBNILP} makes an extensive use of APIs provided by the \textsf{SCIP} framework\autocite{achterberg2008constraint}: \textsf{SCIP} leverages constraint programming (CP) techniques to solve integer programming problems, making it one of the fastest non-commercial MIP solvers in the world. Please see \autoref{fig:ilp_solvers_comparison} for the speed comparison between SCIP and other existing MIP solvers.\footnote{\url{http://scip.zib.de/}} One thing we were curious about during the implementation was whether we could achieve even faster speeds by formulating our problem in a more generic ILP manner and implementing simpler layers between our program and more conventional but still faster solvers such as \textsf{CPLEX} and \textsf{Gurobi}. Also, we wanted \textsf{Bayene} to be more flexibly designed to allow easier adaptation of future developments in our ILP formulation and portability across different solvers.

\subsection{Notes on Branch-and-Cut}
The biggest challenge we faced during the implementation was to modify the solver's branch-and-cut process to our needs. Since we do not specify all the cluster constraints all at once when we transfer our model to the solver, we add them as cutting planes whenever the current relaxed solution violates the conditions. The main problem was that due to the limitations in \textsf{Pyomo} package, we were allowed to add these constraints only when the solver finishes its Branch-and-Cut process completely. This was problematic as this meant that the solver would have to go over branch-and-bound tree to get the integer solution, only to find that it violates the cluster constraint and needs to cut off. Also, the solver will have to completely restart the branch-and-cut tree from scratch, which would add even more time to the solution process. We have tried adjusting several parameters of the solver regarding Branch-and-Cut such as limiting the time spent or number of nodes explored, but these tunings were eventually abandoned as they fail to gurantee any optimality and often terminated the process too early without returning any feasible solution.

Even with the direct interface to the solvers however, there are some issues with the way the solvers handle user constraints. They distinguish two type of cuts users can add to the model, one being \emph{user cuts} and another being \emph{lazy constraints}. User cuts refer to the cutting planes that are implied by the model but cannot be directly inferred by the solver. These constraints tightens the feasible region of the LP relaxation but does not cut off any of the IP region. On the other hand, lazy constraints are the ones that are actually required to get the correct solution but cannot be added all at once because there are too many of them or simply impossible to specify them all in the beginning.

Our cluster constraints fall into the second category - lazy constraints. The problem is that these lazy constraints can only be added to the solver when they reaches \emph{integer feasible} solution, which can take significantly more time than checking lazy constraints at non-integer solution node. Reference manuals of the solvers do not fully specify the reasoning behind this, but our guess is that adding violated cutting planes at every nodes of BnB tree might complicate the solution process too much with excessive branching.

We have eventually settled on the \textsf{Pyomo}-based implementation due to time constraint. We also attempted implementing branch-and-cut outside the solver and control the solution process by ourselves, but this showed to be extremely difficult as branchings occurred very often early in the process and number of nodes on BnB tree grew rapidly beyond our memory control despite different branching strategies we tried.

Understanding various behaviours and techniques of ILP solvers and adjusting them for the best performance on our specific problem requires more thorough investigation on their own, and call for more attention in the future research. 

\section{Experiments}
We experimented with \textsf{Bayene} to see how they perform in practice using reference datasets of different conditions. We also examined how turning on and off \textsf{Bayene}'s different features - cycle cuts, sink-finding heuristic, and gomory fractional cuts - changes its behaviour or performance.

\subsection{Setup}

\begin{itemize}
\item We used pre-calculated BDeu local score files provided by Cussens, available on his \textsf{GOBNILP} website. These score files are based on the reference datasets used for benchmarks in past Bayesian Network literatures, which the original versions can be obtained from the website \textsf{Bayesian Network Repository} by Marco Scutari\footnote{\url{http://www.bnlearn.com/bnrepository/}} with the orignal source information.
\item Each dataset was sampled to have \(100\), \(1000\), and \(10000\) instances, and two scores files were created for every dataset, with parent set size limit of \(2\) and \(3\) respectively.
\item Cussens\autocite{Cussens2015} have indicated that some of the parent candidate sets have been pruned using the methods published by de Campos and Ji.\autocite{de2010properties}
\item Our benchmark have been conducted on Apple Mid 2012 MacBook Air machine with Intel Core i5 3317U 1.7Ghz, 4GB RAM, and OS X 10.11 operating system.
\item While \textsf{Bayene} works on any ILP solver \textsf{Pyomo} supports, we used \textsf{Gurobi} as it was one of the fastest commercial solvers without the problem size limit on academic use.
\item We turned off all the general purpose MIP cuts by setting \textsf{Gurobi} parameter \textsf{Cuts} to 0, except for Gomory fractional cuts by setting \textsf{GomoryPasses} to \(20000000\) (unlimited).
\end{itemize}

\subsection{Experiment 1: All Features Turned On}

For the first experiment, we applied both cluster cuts and cycle cuts, along with Gomory fractional cuts by the solver. Please see \autoref{tab:experiment_1_parent_2} and \autoref{tab:experiment_1_parent_3} for the detailed results. We were able to solve most of the ILP problems within 1-hour limit, ranging from less than a second for \textsf{asia} dataset (8 attributes, 118 ILP variables) to over 20 minutes for \textsf{alarm} (37 attributes, 2736 ILP variables).

\begin{figure}[h!]
\centering
\begin{subfigure}[t]{0.49\textwidth}
\includegraphics[width=\textwidth]{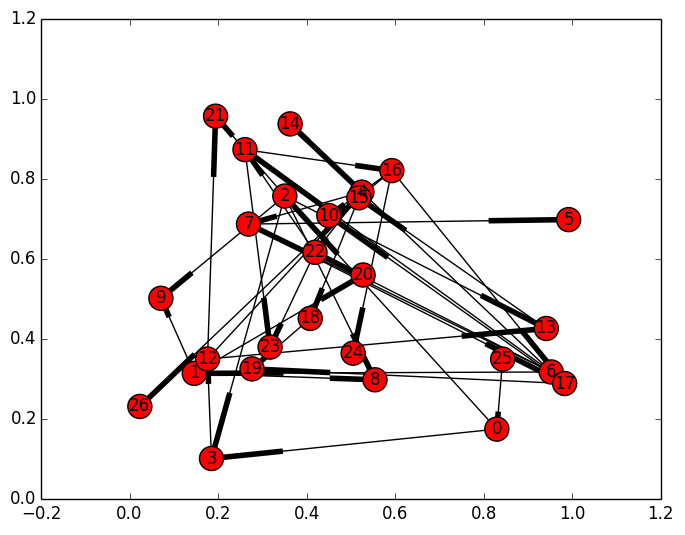}
\caption{Resulting BN structure}
\label{fig:insurance_1000_1_2_graph}
\end{subfigure}
\begin{subfigure}[t]{0.49\textwidth}
\includegraphics[width=\textwidth]{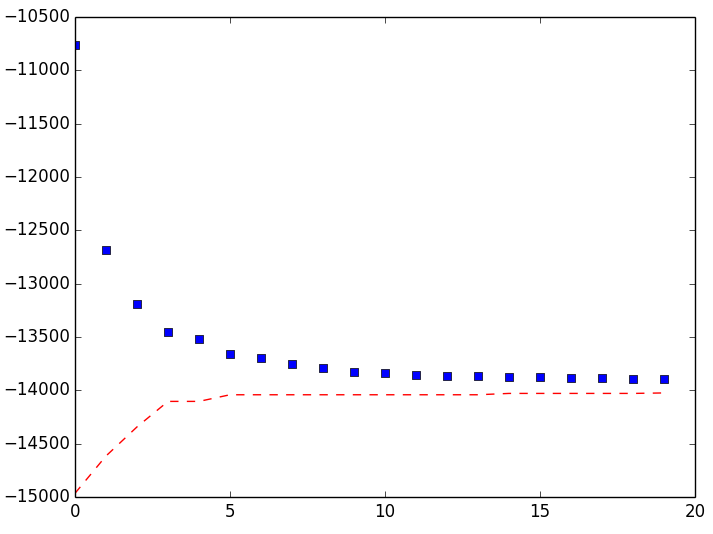}
\caption{Progression of Objective Values}
\label{fig:insurance_1000_1_2_progress}
\end{subfigure}
\caption{Results from \textsf{insurance} with 1000 instances, parent size limit \(=2\).}
\end{figure}

Please see \autoref{fig:insurance_1000_1_2_graph} for example BN structure generated by \textsf{Bayene} for the \textsf{water} dataset. \autoref{fig:insurance_1000_1_2_progress} shows how objective values have progressed during the solution process. Each dots are plotted whenever we add cluster cuts and the solver returns the current solution. ILP objectives are shown as blue dots, and sink heuristic objectives as red dotted lines. Note that sink heuristic objective value do not get changed unless we get bigger objective value.

In the beginning without most of cluster and cycle constraints, we begin with a quite high objective value, but falls rapidly on the next two iterations. The objective value then changes very little until we reach the optimum solution. Interestingly enough, the objective value of sink heuristic reaches near the range of the actual optimum in the very beginning and do not change until the end.

Although cluster cuts and cycle cuts have been invoked same number of times for almost all cases, the total number of cycles that have been ruled out through cycle cuts exceeds the number of cluster cuts by a huge margin. These large number of cuts allow the solver to reach the range of valid solutions more quickly and eventually the optimum.

\subsection{Experiment 2: Without Cycle Cuts}
\begin{figure}[h!]
\centering
\begin{subfigure}[t]{0.49\textwidth}
\includegraphics[width=\textwidth]{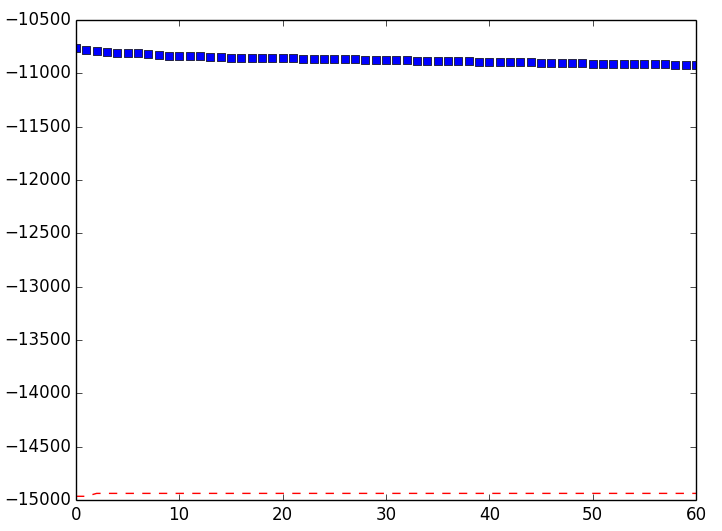}
\caption{\textsf{insurance} with 1000 instances}
\label{fig:insurance_1000_1_2_progress_2}
\end{subfigure}
\begin{subfigure}[t]{0.49\textwidth}
\includegraphics[width=\textwidth]{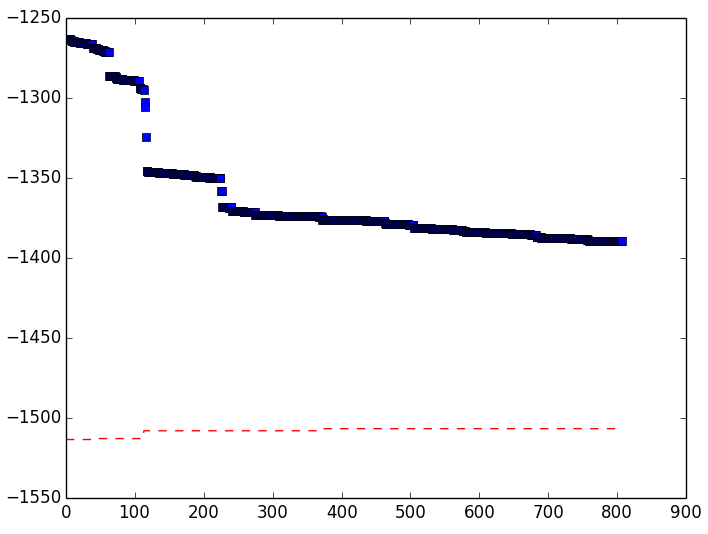}
\caption{\textsf{water} with 100 instances}
\label{fig:water_100_1_2_progress_2}
\end{subfigure}
\caption{Progression of Objective Values from Different Datasets in the Second Experiment. Both problems did not reach optimal solution within the time limit.}
\end{figure}

For the second experiment, we applied just the cluster cuts and kept the Gomory cuts. Please see \autoref{tab:experiment_2_parent_2} and \autoref{tab:experiment_2_parent_3} for the detailed results. We were not able to solve most of the problems within the time limit, as the objective value progressed really slowly as seen from \autoref{fig:insurance_1000_1_2_progress_2} and \autoref{fig:water_100_1_2_progress_2}. Moreover, we observed that the speed of the solver started to slow down over each iteration, while still having the wide gap between the sink heuristic objective value. It seems that adding few cluster cuts already complicates the shape of the feasible region heavily, while they don't cut enough to reach the area with valid solutions. Things get worse since we are restarting the branch-and-bound tree every time we add cluster cuts. We could see that cycle cuts we add by detecting all the elementary cycles serve a significant role in getting the optimal solutions in a reasonable amount of time.

\subsection{Experiment 3: Without Gomory Cuts}
\begin{figure}[h!]
\centering
\begin{subfigure}[t]{0.49\textwidth}
\includegraphics[width=\textwidth]{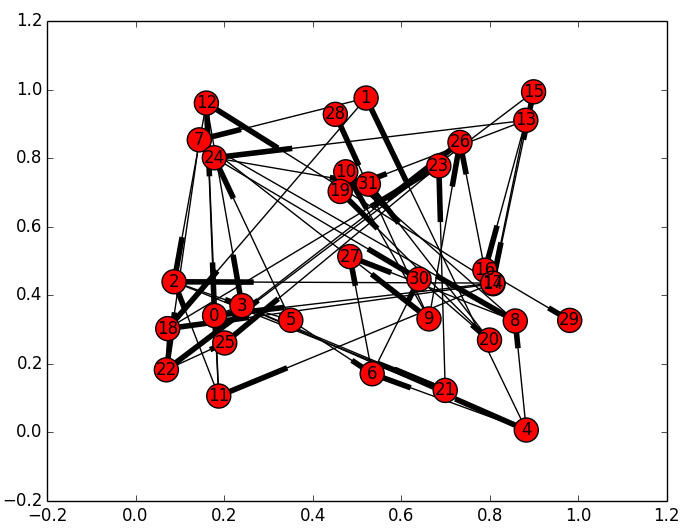}
\caption{Resulting BN structure}
\label{fig:insurance_1000_1_2_graph_3}
\end{subfigure}
\begin{subfigure}[t]{0.49\textwidth}
\includegraphics[width=\textwidth]{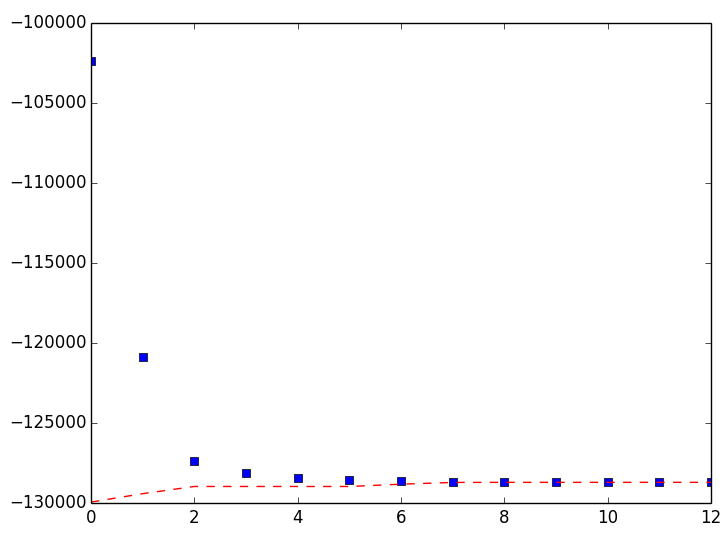}
\caption{Progression of Objective Values}
\label{fig:insurance_1000_1_2_progress_3}
\end{subfigure}
\caption{Results from \textsf{water} with 10000 instances, parent size limit \(=3\).}
\end{figure}

For the third experiment, we applied both cluster cuts and cycle cuts, but completely disabled Gomory frational cuts from the solver. Please see \autoref{tab:experiment_3_parent_2} and \autoref{tab:experiment_3_parent_3} for the detailed results. While the third experiment was able to solve most of the problems as the first experiment did, disabling Gomory cuts improved the solution time significantly in many cases, especially the hardest ones in the first experiment where it took 22 minutes to solve \textsf{alarm} with 10000 instances but 12 minutes in the third experiment. Instead, a little more cluster and cycle cuts were added to solve the problems than the first experiment. This implies that general purpose cutting planes like Gomory cuts that are used to tighten the bound on BnB tree can be countereffective in our cases. Patterns of objective value changes were not different from the first experiment.

\newgeometry{margin=1cm}
\begin{landscape}
\thispagestyle{empty}
\begin{table}
\begin{adjustbox}{width=\linewidth}
  \begin{tabular}{ | l | l | l | l | l | l | l | l | l |}
    \hline
    \textbf{Title} & \textbf{\# Attributes} & \textbf{\# Instances} & \textbf{\# ILP Variables} & \textbf{BDeu Score} & \textbf{Time Elapsed (in sec)} & \textbf{\# Cluster Cut Iterations} & \textbf{\# Cycle Cut Iterations} & \textbf{\# Cycle Cut Count} \\ \hline
    \textsf{asia} & \(8\) & \(100\) & \(41\) & \(-245.644264\) & \(0.1605529785\) & \(3\) & \(3\) & \(11\) \\ \hline
    \textsf{asia} & \(8\) & \(1000\) & \(88\) & \(-2317.411506\) & \(0.493724823\) & \(8\) & \(8\) & \(25\) \\ \hline
    \textsf{asia} & \(8\) & \(10000\) & \(118\) & \(-22466.396546\) & \(0.5763838291\) & \(8\) & \(8\) & \(28\) \\ \Xhline{5\arrayrulewidth}
    \textsf{insurance} & \(27\) & \(100\) & \(266\) & \(-1687.683853\) & \(1.1246800423\) & \(9\) & \(9\) & \(59\) \\ \hline
    \textsf{insurance} & \(27\) & \(1000\) & \(702\) & \(-13892.798172\) & \(31.290997982\) & \(20\) & \(20\) & \(295\) \\ \hline
    \textsf{insurance} & \(27\) & \(10000\) & \(2082\) & \(-133111.964488\) & \(531.714365005\) & \(28\) & \(28\) & \(780\) \\ \Xhline{5\arrayrulewidth}
    \textsf{water} & \(32\) & \(100\) & \(356\) & \(-1501.644722\) & \(3.298566103\) & \(17\) & \(17\) & \(147\) \\ \hline
    \textsf{water} & \(32\) & \(1000\) & \(507\) & \(-13263.115737\) & \(5.2235310078\) & \(16\) & \(16\) & \(227\) \\ \hline
    \textsf{water} & \(32\) & \(10000\) & \(813\) & \(-128810.974528\) & \(10.9444692135\) & \(16\) & \(16\) & \(248\) \\ \Xhline{5\arrayrulewidth}
    \textsf{alarm} & \(37\) & \(100\) & \(591\) & \(-1362.995568\) & \(33.3234539032\) & \(23\) & \(23\) & \(446\) \\ \hline
    \textsf{alarm} & \(37\) & \(1000\) & \(1309\) & \(-11248.39992\) & \(258.938903093\) & \(46\) & \(46\) & \(532\) \\ \hline
    \textsf{alarm} & \(37\) & \(10000\) & \(2736\) & \(-105486.499123\) & \(1331.78003311\) & \(44\) & \(44\) & \(885\) \\ \Xhline{5\arrayrulewidth}
    \textsf{hailfinder} & \(56\) & \(100\) & \(214\) & \(-6021.269394\) & \(1.3409891129\) & \(10\) & \(10\) & \(63\) \\ \hline
    \textsf{hailfinder} & \(56\) & \(1000\) & \(671\) & \(-52473.926982\) & \(10.9388239384\) & \(27\) & \(27\) & \(194\) \\ \hline
    \textsf{hailfinder} & \(56\) & \(10000\) & \(2260\) & \(-498383.409915\) & \(942.131913185\) & \(68\) & \(69\) & \(639\) \\ \Xhline{5\arrayrulewidth}
    \textsf{carpo} & \(60\) & \(100\) & \(2139\) & \(--\) & \(--\) & \(--\) & \(--\) & \(--\) \\ \hline
    \textsf{carpo} & \(60\) & \(1000\) & \(2208\) & \(--\) & \(--\) & \(--\) & \(--\) & \(--\) \\ \hline
    \textsf{carpo} & \(60\) & \(10000\) & \(4354\) & \(--\) & \(--\) & \(--\) & \(--\) & \(--\) \\ \hline
  \end{tabular}
\end{adjustbox}
  \caption{Experiment 1 with parent set size limit \(= 2\). `\(--\)' indicates that the problem was not solved within 1 hour.}
  \label{tab:experiment_1_parent_2}
\end{table}

\begin{table}
  \begin{adjustbox}{width=\linewidth}
  \begin{tabular}{ | l | l | l | l | l | l | l | l | l |}
    \hline
    \textbf{Title} & \textbf{\# Attributes} & \textbf{\# Instances} & \textbf{\# ILP Variables} & \textbf{BDeu Score} & \textbf{Time Elapsed (in sec)} & \textbf{\# Cluster Cut Iterations} & \textbf{\# Cycle Cut Iterations} & \textbf{\# Cycle Cut Count} \\ \hline
    \textsf{asia} & \(8\) & \(100\) & \(41\) & \(-245.644264\) & \(0.2108130455\) & \(4\) & \(4\) & \(11\) \\ \hline
    \textsf{asia} & \(8\) & \(1000\) & \(107\) & \(-2317.411506\) & \(0.4743950367\) & \(7\) & \(7\) & \(27\) \\ \hline
    \textsf{asia} & \(8\) & \(10000\) & \(161\) & \(-22466.396548\) & \(1.2923400402\) & \(10\) & \(10\) & \(41\) \\ \Xhline{5\arrayrulewidth}
    \textsf{insurance} & \(27\) & \(100\) & \(279\) & \(-1686.225878\) & \(1.3009831905\) & \(10\) & \(10\) & \(66\) \\ \hline
    \textsf{insurance} & \(27\) & \(1000\) & \(774\) & \(-13887.350147\) & \(30.3841409683\) & \(17\) & \(17\) & \(269\) \\ \hline
    \textsf{insurance} & \(27\) & \(10000\) & \(3652\) & \(--\) & \(--\) & \(--\) & \(--\) & \(--\) \\ \Xhline{5\arrayrulewidth}
    \textsf{water} & \(32\) & \(100\) & \(482\) & \(-1500.988471\) & \(8.4975321293\) & \(15\) & \(15\) & \(228\) \\ \hline
    \textsf{water} & \(32\) & \(1000\) & \(573\) & \(-13262.367639\) & \(4.0370209217\) & \(12\) & \(12\) & \(210\) \\ \hline
    \textsf{water} & \(32\) & \(10000\) & \(961\) & \(-128705.656236\) & \(29.74208498\) & \(16\) & \(16\) & \(399\) \\ \Xhline{5\arrayrulewidth}
    \textsf{alarm} & \(37\) & \(100\) & \(907\) & \(-1349.227422\) & \(132.99851799\) & \(19\) & \(19\) & \(871\) \\ \hline
    \textsf{alarm} & \(37\) & \(1000\) & \(1928\) & \(-11240.347094\) & \(947.912580967\) & \(47\) & \(47\) & \(841\) \\ \hline
    \textsf{alarm} & \(37\) & \(10000\) & \(2736\) & \(--\) & \(--\) & \(--\) & \(--\) & \(--\) \\ \Xhline{5\arrayrulewidth}
    \textsf{hailfinder} & \(56\) & \(100\) & \(244\) & \(-6019.469926\) & \(1.100315094\) & \(7\) & \(7\) & \(78\) \\ \hline
    \textsf{hailfinder} & \(56\) & \(1000\) & \(761\) & \(-52473.24561\) & \(21.7998039722\) & \(22\) & \(22\) & \(247\) \\ \hline
    \textsf{hailfinder} & \(56\) & \(10000\) & \(3768\) & \(--\) & \(--\) & \(--\) & \(--\) & \(--\) \\ \Xhline{5\arrayrulewidth}
    \textsf{carpo} & \(60\) & \(100\) & \(2139\) & \(--\) & \(--\) & \(--\) & \(--\) & \(--\) \\ \hline
    \textsf{carpo} & \(60\) & \(1000\) & \(2208\) & \(--\) & \(--\) & \(--\) & \(--\) & \(--\) \\ \hline
    \textsf{carpo} & \(60\) & \(10000\) & \(4354\) & \(--\) & \(--\) & \(--\) & \(--\) & \(--\) \\ \hline
  \end{tabular}
\end{adjustbox}
  \caption{Experiment 1 with parent set size limit \(= 3\). `\(--\)' indicates that the problem was not solved within 1 hour.}
  \label{tab:experiment_1_parent_3}
\end{table}
\clearpage
\end{landscape}
\restoregeometry

\newgeometry{margin=1cm}
\begin{landscape}
\thispagestyle{empty}
\begin{table}
\begin{adjustbox}{width=\linewidth}
  \begin{tabular}{ | l | l | l | l | l | l | l | l | l |}
    \hline
    \textbf{Title} & \textbf{\# Attributes} & \textbf{\# Instances} & \textbf{\# ILP Variables} & \textbf{BDeu Score} & \textbf{Time Elapsed (in sec)} & \textbf{\# Cluster Cut Iterations} & \textbf{\# Cycle Cut Iterations} & \textbf{\# Cycle Cut Count} \\ \hline
    \textsf{asia} & \(8\) & \(100\) & \(41\) & \(-245.644264\) & \(1.2105300427\) & \(30\) & \(NA\) & \(NA\) \\ \hline
    \textsf{asia} & \(8\) & \(1000\) & \(88\) & \(-2317.411506\) & \(3.3687419891\) & \(60\) & \(NA\) & \(NA\) \\ \hline
    \textsf{asia} & \(8\) & \(10000\) & \(118\) & \(-22466.396546\) & \(4.5231909752\) & \(64\) & \(NA\) & \(NA\) \\ \Xhline{5\arrayrulewidth}
    \textsf{insurance} & \(27\) & \(100\) & \(266\) & \(--\) & \(--\) & \(--\) & \(NA\) & \(NA\) \\ \hline
    \textsf{insurance} & \(27\) & \(1000\) & \(702\) & \(--\) & \(--\) & \(--\) & \(NA\) & \(NA\) \\ \hline
    \textsf{insurance} & \(27\) & \(10000\) & \(2082\) & \(--\) & \(--\) & \(--\) & \(NA\) & \(NA\) \\ \Xhline{5\arrayrulewidth}
    \textsf{water} & \(32\) & \(100\) & \(356\) & \(--\) & \(--\) & \(--\) & \(NA\) & \(NA\) \\ \hline
    \textsf{water} & \(32\) & \(1000\) & \(507\) & \(--\) & \(--\) & \(--\) & \(NA\) & \(NA\) \\ \hline
    \textsf{water} & \(32\) & \(10000\) & \(813\) & \(--\) & \(--\) & \(--\) & \(NA\) & \(NA\) \\ \Xhline{5\arrayrulewidth}
    \textsf{alarm} & \(37\) & \(100\) & \(591\) & \(--\) & \(--\) & \(--\) & \(NA\) & \(NA\) \\ \hline
    \textsf{alarm} & \(37\) & \(1000\) & \(1309\) & \(--\) & \(--\) & \(--\) & \(NA\) & \(NA\) \\ \hline
    \textsf{alarm} & \(37\) & \(10000\) & \(2736\) & \(--\) & \(--\) & \(--\) & \(NA\) & \(NA\) \\ \Xhline{5\arrayrulewidth}
    \textsf{hailfinder} & \(56\) & \(100\) & \(214\) & \(--\) & \(--\) & \(--\) & \(NA\) & \(NA\) \\ \hline
    \textsf{hailfinder} & \(56\) & \(1000\) & \(671\) & \(--\) & \(--\) & \(--\) & \(NA\) & \(NA\) \\ \hline
    \textsf{hailfinder} & \(56\) & \(10000\) & \(2260\) & \(--\) & \(--\) & \(--\) & \(NA\) & \(NA\) \\ \Xhline{5\arrayrulewidth}
    \textsf{carpo} & \(60\) & \(100\) & \(2139\) & \(--\) & \(--\) & \(--\) & \(NA\) & \(NA\) \\ \hline
    \textsf{carpo} & \(60\) & \(1000\) & \(2208\) & \(--\) & \(--\) & \(--\) & \(NA\) & \(NA\) \\ \hline
    \textsf{carpo} & \(60\) & \(10000\) & \(4354\) & \(--\) & \(--\) & \(--\) & \(NA\) & \(NA\) \\ \hline
  \end{tabular}
\end{adjustbox}
  \caption{Experiment 2 with parent set size limit \(= 2\). `\(--\)' indicates that the problem was not solved within 1 hour.}
  \label{tab:experiment_2_parent_2}
\end{table}

\begin{table}
\begin{adjustbox}{width=\linewidth}
  \begin{tabular}{ | l | l | l | l | l | l | l | l | l |}
    \hline
    \textbf{Title} & \textbf{\# Attributes} & \textbf{\# Instances} & \textbf{\# ILP Variables} & \textbf{BDeu Score} & \textbf{Time Elapsed (in sec)} & \textbf{\# Cluster Cut Iterations} & \textbf{\# Cycle Cut Iterations} & \textbf{\# Cycle Cut Count} \\ \hline
    \textsf{asia} & \(8\) & \(100\) & \(41\) & \(-245.644264\) & \(1.2677178383\) & \(29\) & \(NA\) & \(NA\) \\ \hline
    \textsf{asia} & \(8\) & \(1000\) & \(107\) & \(-2317.411506\) & \(2.9705760479\) & \(50\) & \(NA\) & \(NA\) \\ \hline
    \textsf{asia} & \(8\) & \(10000\) & \(161\) & \(-22466.396547\) & \(5.4340219498\) & \(59\) & \(NA\) & \(NA\) \\ \Xhline{5\arrayrulewidth}
    \textsf{insurance} & \(27\) & \(100\) & \(279\) & \(--\) & \(--\) & \(--\) & \(NA\) & \(NA\) \\ \hline
    \textsf{insurance} & \(27\) & \(1000\) & \(774\) & \(--\) & \(--\) & \(--\) & \(NA\) & \(NA\) \\ \hline
    \textsf{insurance} & \(27\) & \(10000\) & \(3652\) & \(--\) & \(--\) & \(--\) & \(NA\) & \(NA\) \\ \Xhline{5\arrayrulewidth}
    \textsf{water} & \(32\) & \(100\) & \(482\) & \(--\) & \(--\) & \(--\) & \(NA\) & \(NA\) \\ \hline
    \textsf{water} & \(32\) & \(1000\) & \(573\) & \(--\) & \(--\) & \(--\) & \(NA\) & \(NA\) \\ \hline
    \textsf{water} & \(32\) & \(10000\) & \(961\) & \(--\) & \(--\) & \(--\) & \(NA\) & \(NA\) \\ \Xhline{5\arrayrulewidth}
    \textsf{alarm} & \(37\) & \(100\) & \(907\) & \(--\) & \(--\) & \(--\) & \(NA\) & \(NA\) \\ \hline
    \textsf{alarm} & \(37\) & \(1000\) & \(1928\) & \(--\) & \(--\) & \(--\) & \(NA\) & \(NA\) \\ \hline
    \textsf{alarm} & \(37\) & \(10000\) & \(2736\) & \(--\) & \(--\) & \(--\) & \(NA\) & \(NA\) \\ \Xhline{5\arrayrulewidth}
    \textsf{hailfinder} & \(56\) & \(100\) & \(244\) & \(--\) & \(--\) & \(--\) & \(NA\) & \(NA\) \\ \hline
    \textsf{hailfinder} & \(56\) & \(1000\) & \(761\) & \(--\) & \(--\) & \(--\) & \(NA\) & \(NA\) \\ \hline
    \textsf{hailfinder} & \(56\) & \(10000\) & \(3768\) & \(--\) & \(--\) & \(--\) & \(NA\) & \(NA\) \\ \Xhline{5\arrayrulewidth}
    \textsf{carpo} & \(60\) & \(100\) & \(5068\) & \(--\) & \(--\) & \(--\) & \(NA\) & \(NA\) \\ \hline
    \textsf{carpo} & \(60\) & \(1000\) & \(3827\) & \(--\) & \(--\) & \(--\) & \(NA\) & \(NA\) \\ \hline
    \textsf{carpo} & \(60\) & \(10000\) & \(16391\) & \(--\) & \(--\) & \(--\) & \(NA\) & \(NA\) \\ \hline
  \end{tabular}
\end{adjustbox}
  \caption{Experiment 2 with parent set size limit \(= 3\). `\(--\)' indicates that the problem was not solved within 1 hour.}
  \label{tab:experiment_2_parent_3}
\end{table}
\clearpage
\end{landscape}
\restoregeometry

\newgeometry{margin=1cm}
\begin{landscape}
\thispagestyle{empty}
\begin{table}
\begin{adjustbox}{width=\linewidth}
  \begin{tabular}{ | l | l | l | l | l | l | l | l | l |}
    \hline
    \textbf{Title} & \textbf{\# Attributes} & \textbf{\# Instances} & \textbf{\# ILP Variables} & \textbf{BDeu Score} & \textbf{Time Elapsed (in sec)} & \textbf{\# Cluster Cut Iterations} & \textbf{\# Cycle Cut Iterations} & \textbf{\# Cycle Cut Count} \\ \hline
    \textsf{asia} & \(8\) & \(100\) & \(41\) & \(-245.644264\) & \(0.229927063\) & \(4\) & \(4\) & \(11\) \\ \hline
    \textsf{asia} & \(8\) & \(1000\) & \(88\) & \(-2317.411506\) & \(0.5296328068\) & \(8\) & \(8\) & \(25\) \\ \hline
    \textsf{asia} & \(8\) & \(10000\) & \(118\) & \(-22466.396546\) & \(0.6273970604\) & \(8\) & \(8\) & \(28\) \\ \Xhline{5\arrayrulewidth}
    \textsf{insurance} & \(27\) & \(100\) & \(266\) & \(-1687.683853\) & \(1.0839219093\) & \(9\) & \(9\) & \(59\) \\ \hline
    \textsf{insurance} & \(27\) & \(1000\) & \(702\) & \(-13892.798172\) & \(26.9869270325\) & \(21\) & \(21\) & \(318\) \\ \hline
    \textsf{insurance} & \(27\) & \(10000\) & \(2082\) & \(-133111.964488\) & \(319.534489155\) & \(28\) & \(28\) & \(776\) \\ \Xhline{5\arrayrulewidth}
    \textsf{water} & \(32\) & \(100\) & \(356\) & \(-1501.644722\) & \(6.8530170918\) & \(20\) & \(20\) & \(183\) \\ \hline
    \textsf{water} & \(32\) & \(1000\) & \(507\) & \(-13263.115737\) & \(4.921047926\) & \(16\) & \(16\) & \(186\) \\ \hline
    \textsf{water} & \(32\) & \(10000\) & \(813\) & \(-128810.974528\) & \(48.8902380466\) & \(19\) & \(19\) & \(981\) \\ \Xhline{5\arrayrulewidth}
    \textsf{alarm} & \(37\) & \(100\) & \(591\) & \(-1362.995568\) & \(28.4560148716\) & \(23\) & \(23\) & \(434\) \\ \hline
    \textsf{alarm} & \(37\) & \(1000\) & \(1309\) & \(-11248.39992\) & \(173.517156839\) & \(47\) & \(47\) & \(548\) \\ \hline
    \textsf{alarm} & \(37\) & \(10000\) & \(2736\) & \(-105486.499123\) & \(721.979949951\) & \(45\) & \(45\) & \(869\) \\ \Xhline{5\arrayrulewidth}
    \textsf{hailfinder} & \(56\) & \(100\) & \(214\) & \(-6021.269394\) & \(1.3437559605\) & \(10\) & \(10\) & \(63\) \\ \hline
    \textsf{hailfinder} & \(56\) & \(1000\) & \(671\) & \(-52473.926982\) & \(9.6990509033\) & \(24\) & \(24\) & \(191\) \\ \hline
    \textsf{hailfinder} & \(56\) & \(10000\) & \(2260\) & \(--\) & \(--\) & \(--\) & \(--\) & \(--\) \\ \Xhline{5\arrayrulewidth}
    \textsf{carpo} & \(60\) & \(100\) & \(2139\) & \(--\) & \(--\) & \(--\) & \(--\) & \(--\) \\ \hline
    \textsf{carpo} & \(60\) & \(1000\) & \(2208\) & \(--\) & \(--\) & \(--\) & \(--\) & \(--\) \\ \hline
    \textsf{carpo} & \(60\) & \(10000\) & \(4354\) & \(--\) & \(--\) & \(--\) & \(--\) & \(--\) \\ \hline
  \end{tabular}
\end{adjustbox}
  \caption{Experiment 3 with parent set size limit \(= 2\). `\(--\)' indicates that the problem was not solved within 1 hour.}
  \label{tab:experiment_3_parent_2}
\end{table}

\begin{table}
\begin{adjustbox}{width=\linewidth}
  \begin{tabular}{ | l | l | l | l | l | l | l | l | l |}
    \hline
    \textbf{Title} & \textbf{\# Attributes} & \textbf{\# Instances} & \textbf{\# ILP Variables} & \textbf{BDeu Score} & \textbf{Time Elapsed (in sec)} & \textbf{\# Cluster Cut Iterations} & \textbf{\# Cycle Cut Iterations} & \textbf{\# Cycle Cut Count} \\ \hline
    \textsf{asia} & \(8\) & \(100\) & \(41\) & \(-245.644264\) & \(0.2282519341\) & \(4\) & \(4\) & \(11\) \\ \hline
    \textsf{asia} & \(8\) & \(1000\) & \(107\) & \(-2317.411506\) & \(0.5901920795\) & \(7\) & \(7\) & \(27\) \\ \hline
    \textsf{asia} & \(8\) & \(10000\) & \(161\) & \(-22466.396546\) & \(1.231169939\) & \(10\) & \(10\) & \(41\) \\ \Xhline{5\arrayrulewidth}
    \textsf{insurance} & \(27\) & \(100\) & \(279\) & \(-1686.225878\) & \(1.3578009605\) & \(10\) & \(10\) & \(66\) \\ \hline
    \textsf{insurance} & \(27\) & \(1000\) & \(774\) & \(-13887.350147\) & \(20.3226950169\) & \(17\) & \(17\) & \(269\) \\ \hline
    \textsf{insurance} & \(27\) & \(10000\) & \(3652\) & \(--\) & \(--\) & \(--\) & \(--\) & \(--\) \\ \Xhline{5\arrayrulewidth}
    \textsf{water} & \(32\) & \(100\) & \(482\) & \(-1500.968391\) & \(8.6199500561\) & \(13\) & \(13\) & \(206\) \\ \hline
    \textsf{water} & \(32\) & \(1000\) & \(573\) & \(-13262.465272\) & \(7.5233428478\) & \(13\) & \(13\) & \(227\) \\ \hline
    \textsf{water} & \(32\) & \(10000\) & \(961\) & \(-128705.731312\) & \(28.2011299133\) & \(13\) & \(13\) & \(456\) \\ \Xhline{5\arrayrulewidth}
    \textsf{alarm} & \(37\) & \(100\) & \(907\) & \(-1349.227422\) & \(121.306695938\) & \(19\) & \(19\) & \(871\) \\ \hline
    \textsf{alarm} & \(37\) & \(1000\) & \(1928\) & \(-11240.347094\) & \(754.878758907\) & \(44\) & \(44\) & \(854\) \\ \hline
    \textsf{alarm} & \(37\) & \(10000\) & \(6473\) & \(--\) & \(--\) & \(--\) & \(--\) & \(--\) \\ \Xhline{5\arrayrulewidth}
    \textsf{hailfinder} & \(56\) & \(100\) & \(244\) & \(-6019.469926\) & \(1.1171729565\) & \(7\) & \(7\) & \(78\) \\ \hline
    \textsf{hailfinder} & \(56\) & \(1000\) & \(761\) & \(-52473.24561\) & \(14.2832479477\) & \(22\) & \(22\) & \(258\) \\ \hline
    \textsf{hailfinder} & \(56\) & \(10000\) & \(3768\) & \(--\) & \(--\) & \(--\) & \(--\) & \(--\) \\ \Xhline{5\arrayrulewidth}
    \textsf{carpo} & \(60\) & \(100\) & \(5068\) & \(--\) & \(--\) & \(--\) & \(--\) & \(--\) \\ \hline
    \textsf{carpo} & \(60\) & \(1000\) & \(3827\) & \(--\) & \(--\) & \(--\) & \(--\) & \(--\) \\ \hline
    \textsf{carpo} & \(60\) & \(10000\) & \(16391\) & \(--\) & \(--\) & \(--\) & \(--\) & \(--\) \\ \hline
  \end{tabular}
\end{adjustbox}
  \caption{Experiment 3 with parent set size limit \(= 3\). `\(--\)' indicates that the problem was not solved within 1 hour.}
  \label{tab:experiment_3_parent_3}
\end{table}
\clearpage
\end{landscape}
\restoregeometry

\subsection{Performance of Sink Finding Heuristic}
\begin{figure}[h]
\centering
\includegraphics[width=\textwidth]{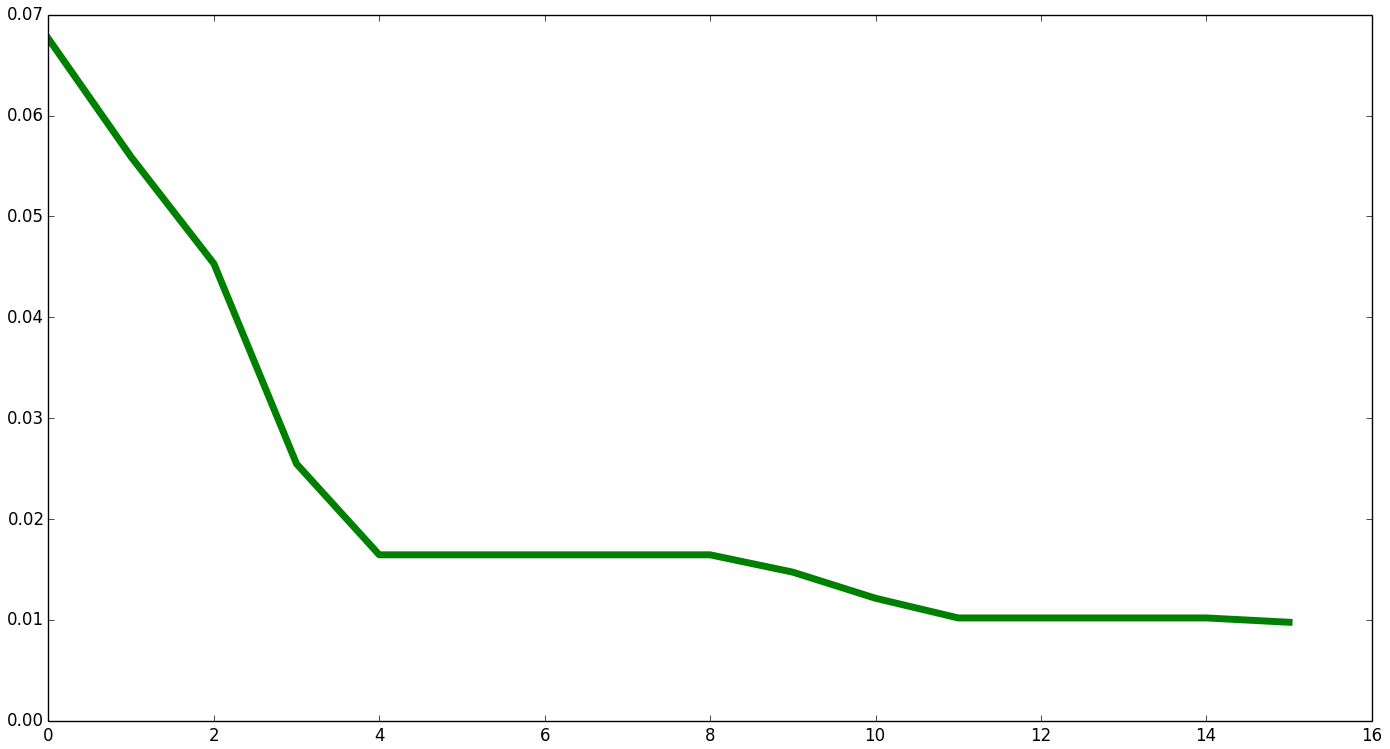}
\caption{Percentage difference between best heuristic objective at each iteration and the final optimal objective value, on \textsf{insurance} dataset of 1000 instances and parent size limit \(=3\).}
\label{fig:sink_typical_gap}
\end{figure}

One interesting aspect of our sink finding heuristic is the proximity of its output to the final optimal solution. For the case of \textsf{insurance} dataset as seen on \autoref{fig:sink_typical_gap}, the heuristic algorithm was able to produce a solution with an objective value that differed around 6\% from the final optimal objective value, even for the first iteration. It soon reached around 1\% in next few iterations. While we only have limited knowledge about the shape of polytopes for BN structures, we can see that our sink finding heuristic can reach the vicinity of the optimal solution quite well. Although \textsf{Bayene} sends the solutions generated by the heuristic to the solver for \emph{warm starts}, it seems that the solver does not actually make much use of them since the solutions for the relaxed problem yet without all the necessary cuts are bigger than the sink finding heuristic solution. Further study of BN structure polytopes and solutions generated by the sink finding heuristic would help us to get optimal solutions faster.

\chapter{Conclusion}
This dissertation reviewed the conceptual foundation behind Bayesian network and studied formulating the problem of learning the BN structure from data as an integer linear programming problem. We went over the inner workings of score metrics used to measure the statistical fit of the BN structure to the dataset, and presented the ILP formulation based on the decomposability of the score metric. In order to deal with the exponential number of constraints, we investigated different ways to add constraints to the model on the fly as cutting planes rather than fully specifying them initially.

We implemented the ILP formulation and cutting planes as a computer software and conducted a benchmark on various reference datasets. We saw that ruling out all the cycles found in the solution at each iteration is critical to reaching the optimality in a reasonable amount of time. We also found out that general purpose cutting planes such as Gomory fractional cuts that are used to tighten the bound on BnB tree can backfire in some cases such as ours. Lastly, we discovered that our sink finding heuristic algorithm returns solutions that are quite close to the final optimal solution very early on in the process.

\section{Future Directions}
\label{sec:future_directions}
This section will present some of the ideas for the further development of \textsf{Bayene} from two aspects: its statistical modelling capabilities and mathematical optimisation techniques employed.

\subsection{Statistical Modelling}
\label{sec:future_stat}

\subsubsection{Alternate Scoring Functions}
For this dissertation, we focused on using Bayesian Dirichlet-based score metric, which is based on the assumption of multinomial data and Dirichlet prior. We used BDeu score metric, which adds additional assumptions of likelihood equivalence and uniform prior probabilities. In addition to Dirichlet-based score metric, there are a number of different information theoretic score functions such as MDL and BIC used in BN literatures. Also, there have been some recent development on score metrics such as SparsityBoost by Brenner and Sontag\autocite{BrennerS13} that reduces computational burden and attempts to incorporate aspects of conditional independence testing. Understanding differences between these score metrics would be important in making the effectiveness of our Bayesian network structure learning as a statistical model.

\subsubsection{Other Types of Statistical Distribution}
Adding on to alternate score metrics, versatility of Bayesian network and its structure learning can be expanded by making it applicable to different types of distribution. There already have been done for learning BN structure on continous distribution, but mostly based on conditional independence information. Figuring out ways to allow more types of distribution, especially for the ILP formulation, would be an interesting and worthwhile challenge.

\subsection{Optimisation}
\label{sec:future_optimisation}

\subsubsection{Leveraging the Graph Structure}
While we did not directly make much use of the fact that the structure of Bayesian network is DAG, there were few researches that did in the last few years, including the one by Studeny et al.\autocite{studeny2014learning} that introduced the concept of characteristic imset, which stems from the property of Markov equivalent structures described in \autoref{sec:key_characteristics}, and another with a set of additional treewidth constraints by Parviainen et al.\autocite{parviainen2014learning}. Empirical results on these approaches showed to be significantly slower than ours, but it would be interesting to go further with these leads from combinatorial optimisation perspective.

\subsubsection{Advanced Modelling Techniques}
We benchmarked on the pre-calculated local score files that have parent set size limit of either 2 or 3. While our formulation theoretically works on bigger parent set sizes, we currently haven't employed more advanced techniques such as column generation that could make it possible to deal with extremely large number of variables. Further development of incorporating such techniques to \textsf{Bayene} would allow us to handle larger datasets.

\subsubsection{Alternate Optimisation Paradigm}
Aside from the integer linear programming, there have been efforts to use alternate optimsation scheme such as constraint programming.\cite{vanmachine} While they were not decisively better than our ILP approach, they did show some promising results. It would be worthwhile to examine the inner workings of their approach in order to improve our formulation.

\subsubsection{Deeper Integration with Branch-and-Cut}
As seen from \autoref{sec:implementation_details}, we had some issues with adjusting the solver's branch-and-cut algorithm to our needs, as there were complications resulting from various techniques and restrictions involved with the solver programs. In order to make \textsf{Bayene} suitable for more learning tasks, thorough inspection of how the ILP solvers perform optimisation would be needed to prevent inefficent operations.

\addappheadtotoc
\appendix
\appendicestocpagenum

\chapter{Software Instructions}
\textsf{Bayene} can be downloaded from the following link: \url{https://link.iamblogger.net/4khv1}. \textsf{Bayene} itself is not a standalone program but rather a \textsf{Python} library, so the user needs to inherit the class from the package to his or her application. For evaluation purposes, we provide a test script file \textsf{sample_script.py} that allows the user to testdrive \textsf{Bayene}.

There's no formal install functionality yet on \textsf{Bayene}, so the user needs to install all the dependencies manually. \textsf{Bayene} was written for \textsf{CPython} 2.7 series, and will not work on any other implementation of \textsf{Python} such as \textsf{Python} 3 or \textsf{PyPy}. Please download the appropriate version of \textsf{Python} 2.7 for your platform from \url{https://www.python.org/downloads/}.

If your system does not have \textsf{pip}, please refer to \url{https://pip.pypa.io/en/latest/installing.html} for install instructions. After installing \textsf{pip}, please turn on \textsf{Command Prompt} on \textsf{Windows} or \textsf{Terminal} on \textsf{OS X} or \textsf{Linux} with administrator or root access and type \textsf{pip install pyomo numpy scipy}, which will install all the required libaries for \textsf{Bayene}. In addition, you need to install \textsf{gurobipy} package included with the installation of \textsf{Gurobi} solver, which the instructions are provided in \url{http://www.gurobi.com/documentation/}.

After installing all the dependencies, please open \textsf{sample_script.py} with a plain text editor. Please edit the string on line 12 to specify the score files that needs to be tested. Please note that all the score files used for this dissertation is available in \url{http://www-users.cs.york.ac.uk/~jc/research/uai11/ua11_scores.tgz}.

Lastly, go back to \textsf{Command Prompt} or \textsf{Terminal}, navigate to the directory where \textsf{sample_script.py} is located and type \textsf{python sample_script.py}. Please refer to the source code for further information.

\newpage
\pagenumbering{gobble}

\printbibliography

\end{document}